\documentclass[11pt,a4paper]{article}
\usepackage{times,latexsym}
\usepackage{url}
\usepackage[T1]{fontenc}
\usepackage{bm}
\usepackage{booktabs} 
\usepackage{multirow,multicol}
\usepackage{colortbl}
\usepackage{kotex}
\usepackage[acceptedWithA,nohyperref]{tacl2018v2}
\usepackage{xspace,mfirstuc,tabulary}

\newcommand{\cg}[1]{\cellcolor{lightgray}#1}
\newcommand{\cgg}[1]{\cellcolor{gray}#1}
\newcommand{\mf}[1]{\multicolumn{2}{c}{\bf #1}}
\newcommand{\smf}[1]{\multicolumn{2}{c}{#1}}

\newif\iftaclinstructions
\taclinstructionsfalse % AUTHORS: do NOT set this to true
\iftaclinstructions
\renewcommand{\confidential}{}
\renewcommand{\anonsubtext}{(No author info supwplied here, for consistency with
TACL-submission anonymization requirements)}
\newcommand{\instr}
\fi

\iftaclpubformat % this "if" is set by the choice of options

\else

\fi

\newcommand\blfootnote[1]{%
  \begingroup
  \renewcommand\thefootnote{}\footnote{#1}%
  \addtocounter{footnote}{-1}%
  \endgroup
}
\newcommand{\STAB}[1]{\begin{tabular}{@{}c@{}}#1\end{tabular}}

\title{Multilingual Denoising Pre-training for Neural Machine Translation}
\author{Yinhan Liu*, Jiatao Gu*, Naman Goyal*, Xian Li, Sergey Edunov\\
\textbf{Marjan Ghazvininejad, Mike Lewis, Luke Zettlemoyer} \\
Facebook AI Research\\
\texttt{\{yinhanliu,jgu,naman,xianl,edunov} \\
\texttt{ghazvini,mikelewis,lsz\}}
\texttt{@fb.com} \\
}
\usepackage{amsmath,amssymb}
\usepackage{natbib}
\usepackage{graphicx}
\usepackage{cleveref}
\usepackage{subfigure}
\usepackage{enumitem}

\crefformat{section}{\S#2#1#3} % see manual of cleveref, section 8.2.1
\crefformat{subsection}{\S#2#1#3}
\crefformat{subsubsection}{\S#2#1#3}
\begin{document}

\maketitle
\
\blfootnote{\textbf{*} Equal contribution.}
\begin{abstract}
This paper demonstrates that multilingual denoising pre-training produces significant performance gains across a wide variety of machine translation (MT) tasks. We present \textit{mBART} -- a sequence-to-sequence denoising auto-encoder pre-trained on large-scale monolingual corpora in many languages using the BART objective~\cite{lewis2019bart}. 
mBART is the first method for pre-training a  complete sequence-to-sequence model by denoising  full texts in multiple languages, while previous approaches have focused only on the encoder, decoder, or reconstructing parts of the text.  %\mike{I think the previous sentence needs work}
Pre-training a complete model allows it to be directly fine tuned for supervised (both sentence-level and document-level) and unsupervised machine translation, with no task-specific modifications. We demonstrate that adding mBART initialization produces performance gains in all but the highest-resource settings, including up to 12 BLEU points for low resource MT and over 5 BLEU points for many document-level and unsupervised models. We also show it also enables new types of transfer to language pairs with no bi-text or that were not in the pre-training corpus, and present extensive analysis of which factors contribute the most to effective pre-training. 
\end{abstract}
\section{Introduction}
\label{sec:intro}
% what is the problem
Despite its wide adoption for other NLP  tasks~\cite{devlin2018bert,liu2019roberta, yang2019xlnet,lewis2019bart,raffel2019exploring}, self-supervised pretraining is not yet common practice in machine translation (MT). Existing MT approaches only pre-train parts of the model, including the encoder~\cite{lample2019cross} and the decoder~\cite{edunov2019pre}, or use pre-training objectives that only reconstruct parts of  text~\cite{song2019mass}, or only focus on English corpora~\cite{lewis2019bart,raffel2019exploring}. In this paper, we show that significant performance gains are possible by pre-training a complete autoregressive model with an objective that noises and reconstructs full texts across many languages. 

In this work, we present \textit{mBART} -- a multilingual sequence-to-sequence (Seq2Seq) denoising auto-encoder.
mBART is trained by applying the BART~\cite{lewis2019bart} to large-scale monolingual corpora across many languages. The input texts are noised by masking phrases and permuting sentences, and a single Transformer~\cite{vaswani2017attention} model is learned to recover the texts. 
Different from other pre-training approaches for MT~\cite{lample2019cross,song2019mass}, mBART pre-trains a complete autoregressive Seq2Seq model.
%\mike{Needs some more caveats}. 
mBART is trained once for all languages, providing a set of parameters that can be fine-tuned for any of the language pairs in both supervised and unsupervised settings, without any task-specific or language-specific modifications or initialization schemes. 

Extensive experiments demonstrate that this simple approach works remarkably well. We first focus on existing MT benchmarks. For supervised sentence-level MT,
mBART initialization leads to significant gains (up to 12 BLEU points) across low/medium-resource pairs (<10M bi-text pairs), without sacrificing performance in high-resource settings. These results further improve with back-translation (BT), setting a new state-of-the-art on WMT16 English-Romanian and the FloRes test sets. For document-level MT, our document-level pre-training improves results by up to 5.5. For the unsupervised case, we see consistent gains and produce the first non-degenerate results for less related language pairs (e.g., 9.5 BLEU gain on Nepali-English). Previous pre-training schemes have only considered subsets of these tasks, but we compare performance where possible and demonstrate that mBART consistently performs the best.

We also show that mBART enables new types of transfer across language pairs. 
For example, fine-tuning on bi-text in one language pair (e.g., Korean-English) creates a model that can translate from all other languages in the monolingual pre-training set (e.g., Italian-English), with no further training. We also show that languages not in pre-training corpora can benefit from mBART, strongly suggesting that the initialization is at least partially language universal. Finally, we present a detailed analysis of which factors contribute the most to effective pre-training, including the number of languages and their overall similarity.  
\begin{figure*}[tbph]
    \centering
    \includegraphics[width=\textwidth]{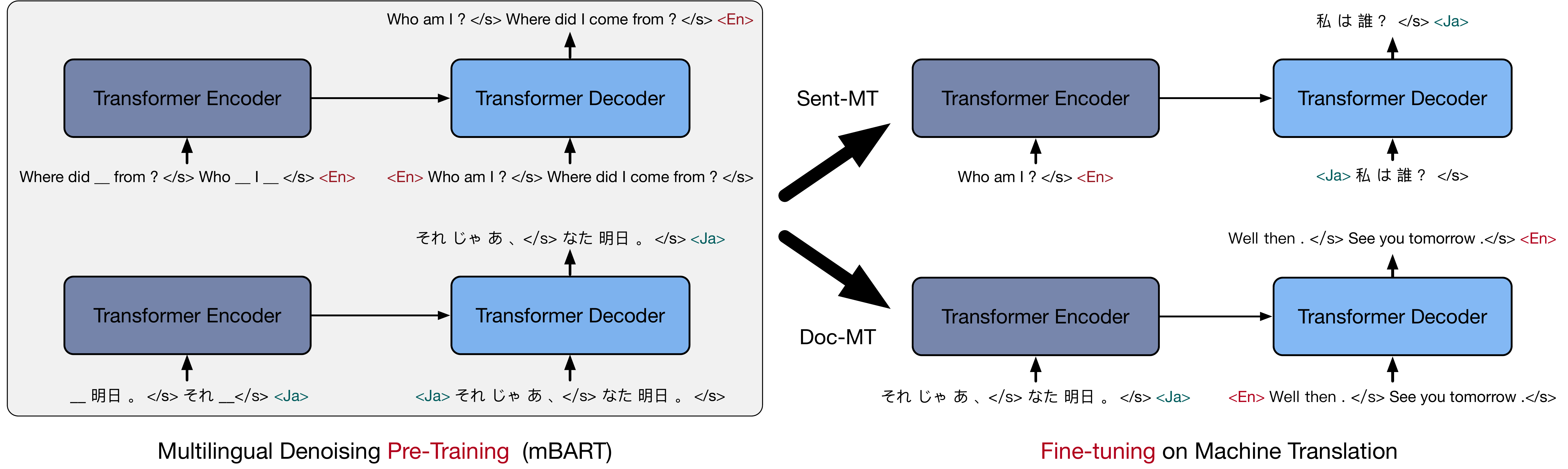}
    \caption{Framework for our Multilingual Denoising Pre-training (left) and fine-tuning on downstream MT tasks (right), where we use (1) sentence permutation (2) word-span masking as the injected noise. A special language id token is added at both the encoder and decoder. One multilingual pre-trained model is used for all tasks. 
    \label{fig:pt:framework}}
\end{figure*}

\section{Multilingual Denoising Pre-training}
\label{sec:model}
%In this section, we introduce the large-scale multilingual corpora used in our pre-training stage, along with the details about the optimization and the resulted models for fine-tuning on translation. Figure~\ref{fig:pt:framework} shows an example of pre-training.
%Here, we describe the data, pre-training procedure, and resultant models used in the rest of our experiments (Figure~\ref{fig:pt:framework}).

We use a large-scale common crawl (CC) corpus (\cref{sec:pt:data}) to pre-train BART models (\cref{sec:pt:mbart}). Our experiments in the later sections involve finetuning a range of models pre-trained on different subsets of the CC languages \cref{sec:pt:models}).

%We concatenate monolingual corpus across different languages and train our pretrain model on it. We apply multiple training strategies during pretraining, and compare these pretraining models under different scenarios. 
\subsection{Data: CC25 corpus}
\label{sec:pt:data}
\paragraph{Datasets}
 %We first collect our pre-training corpus 
 We pre-train on a subset of 25 languages 
 -- CC25 -- extracted from the Common Crawl (CC) ~\citep{wenzek2019ccnet, conneau2019unsupervised}\footnote{\url{https://commoncrawl.org}}. 
 %We select a subset of text in $25$ languages, including different 
 CC25 includes languages from different families and with varied amounts of text (Table~\ref{tab:datastats}). 
Following \citet{lample2019cross}, we re-balanced the corpus by up/down-sampling text from each language $i$ with a ratio $\lambda_i$:
\begin{equation}
    \lambda_i = \frac{1}{p_i}\cdot\frac{p_i^\alpha}{\sum_{i}{p_i^\alpha}},
\end{equation}
where $p_i$ is the percentage of each language in CC-25. We use the smoothing parameter $\alpha=0.7$. 
%During pre-training, we sample instances with a sampling rate proportional to number of instances in monolingual corpus of each language (similarly to \cite{lample2019cross}). This sampling rate is smoothed with exponential smoothing parameter $\alpha=0.7$.  \mike{The last 2 sentences need detail/clarifying}\jgu{I thought it might be too much detail and refer to the original XLM paper.}

\begin{table}[t]
\begin{center}
\small
\begin{tabular}[b]{llrr}
\toprule
\textbf{Code} & 
\textbf{Language} & 
\textbf{Tokens/M} & \textbf{Size/GB}\\
\midrule
{\bf En }& English & 55608 & 300.8 \\
{\bf Ru }& Russian & 23408 & 278.0 \\
{\bf Vi }& Vietnamese & 24757 & 137.3 \\
{\bf Ja }& Japanese & 530 (*) & 69.3 \\
{\bf De}& German & 10297 & 66.6 \\
{\bf Ro }& Romanian & 10354 & 61.4\\
{\bf Fr }& French & 9780 & 56.8 \\
{\bf Fi }& Finnish & 6730 & 54.3 \\
{\bf Ko }& Korean & 5644 & 54.2 \\
{\bf Es }& Spanish & 9374 & 53.3 \\
{\bf Zh } & Chinese (Sim) & 259 (*) & 46.9\\
{\bf It }& Italian & 4983 & 30.2 \\
{\bf Nl }& Dutch & 5025 & 29.3 \\
{\bf Ar }& Arabic & 2869 & 28.0 \\
{\bf Tr }& Turkish & 2736 & 20.9 \\
{\bf Hi }& Hindi & 1715 & 20.2 \\
{\bf Cs }& Czech & 2498 & 16.3 \\
{\bf Lt }& Lithuanian & 1835 & 13.7 \\
{\bf Lv }& Latvian & 1198 & 8.8 \\
{\bf Kk }& Kazakh & 476 & 6.4\\
{\bf Et }& Estonian & 843 & 6.1 \\
{\bf Ne }& Nepali & 237 & 3.8 \\
{\bf Si }& Sinhala & 243 & 3.6 \\
{\bf Gu }& Gujarati & 140 & 1.9\\
{\bf My }& Burmese & 56 & 1.6 \\
\bottomrule
\end{tabular}
\caption{\textbf{Languages and Statistics of the CC25 Corpus.} A list of 25 languages ranked with monolingual corpus size. Throughout this paper, we replace the language names with their ISO codes for simplicity.
(*) Chinese and Japanese corpus are not segmented, so the tokens counts here are sentences counts\label{tab:datastats}}
\end{center}
\end{table}

\paragraph{Pre-processing}
We tokenize with a sentence-piece model~\citep[SPM,][]{kudo-richardson-2018-sentencepiece} learned on the full CC data that includes $250,000$ subword tokens. While not all of these languages are used for pre-training, this tokenization supports fine-tuning on additional languages. We do not apply additional preprocessing, such as true-casing or normalizing punctuation/characters. 
%In the end, we obtain a large dictionary of $250,000$ subword tokens.
%Instead, we directly preprocess
%the raw corpus into subword units with the same sentence-piece model (SPM)~\cite{kudo-richardson-2018-sentencepiece} learned from the full CC corpora. 
%The final dictionary size is $250,000$ sub-word tokens. Note that our dictionary contains symbols that are not seen in our pre-training stage. However, it enables us to extend our models to unseen languages in future work.
% For simplicity, all of our pre-training and fine-tuning models share the same dictionary.
%. We build the dictionary based on SPM on 100 languages, which results in a  dictionary size of 250,000 BPEed tokens. %All of our mBART25, mBART06, and bilingual models share the same dictionary (see section 4.1). 
%\yinhan{todo:Naman: add the up/down sampling tricks here.}

\subsection{Model: mBART}
\label{sec:pt:mbart}

Our models follow the BART~\cite{lewis2019bart} sequence-to-sequence pre-training scheme, as reviewed in this section. 
While BART was only pretrained for English, we systematically study the effects of pre-training on different sets of languages.

%In this work, we built up our pre-training model -- mBART -- based on \citet[BART,][]{lewis2019bart}, a recently proposed sequence-to-sequence pre-training strategy for both generation and classification tasks. 
%The original BART was trained on English corpora only, while we extend the effort in a multilingual scenario with larger scale.

\paragraph{Architecture}
We use a standard sequence-to-sequence Transformer architecture~\cite{vaswani2017attention}, with $12$ layers of encoder and $12$ layers of decoder with model dimension of $1024$ on $16$ heads ($\sim 680$M parameters). We include an additional layer-normalization layer on top of both the encoder and decoder, which we found stabilized training at FP16 precision. 
% \jgu{TODO: do we mention extra-layernorm?}

\paragraph{Learning}
\label{sec:pt:learn}
%Suppose we pretrain the model on 
Our training data covers $K$ languages: $\mathcal{D}=\{\mathcal{D}_1, ..., \mathcal{D}_K \}$ where each $\mathcal{D}_i$ is a collection of monolingual documents in language $i$.
We (1) assume access to a noising function $g$, defined below, that corrupts text, and (2) train the model to predict the original text $X$ given $g(X)$. More formally, we aim to maximize $\mathcal{L}_\theta$: 
\begin{equation}
    \mathcal{L}_\theta = 
    \sum_{\mathcal{D}_i\in \mathcal{D}}
    \sum_{X\in \mathcal{D}_i}\log P(X | g(X); \theta)~,
    \label{eq:learning}
\end{equation}
where $X$ is an instance in language $i$ and the distribution $P$ is defined by the Seq2Seq model. 

\paragraph{Noise function}
Following~\newcite{lewis2019bart}, we use two types of noise in $g$. We first remove spans of text and replace them with a mask token. We mask 35\% of the words in each instance by random sampling a span length according to a Poisson distribution ($\lambda=3.5$).  
We also permute the order of sentences within each instance.
The decoder input is the original text with one position offset. A language id symbol <LID> is used as the initial token to predict the sentence. It is also possible to use other noise types, such as those in \newcite{lample2018phrase}, but we leave the exploration of the optimal noising strategy to future work.

\paragraph{Instance format}
%We first sample a language id symbol <LID> for each batch. 
For each instance of a batch, we sample a language id symbol <LID>, and
we pack as many consecutive sentences as possible sampled from the corresponding corpus of <LID>, until either it hits the document boundary or reaches the $512$ max token length. Sentences in the instance are separated by the end of sentence (</S>) token. Then, we append the selected <LID> token to represent the end of this instance. 
Pre-training at ``multi-sentence'' level enables us to work on both sentence and document translation.
%For pre-training, we also prepend a Start of Sentence (<S>) token to each instance.

\paragraph{Optimization}
% We use the Fairseq~\cite{ott2019fairseq} for all experiments.  
Our full model (including $25$ languages) is trained on 256 Nvidia V100 GPUs (32GB) for 500K steps. 
The total batch size is around $128$K tokens per GPU,
%We set the max-token to be 1600 and update-freq to be 8. 
matching BART~\cite{lewis2019bart} configuration. We use the Adam optimizer ($\epsilon=1\mathrm{e}{-6}$, $\beta_2 = 0.98$) and linear learning rate decay scheduling. The total training time was approximately 2.5 weeks. We started the training with dropout $0.1$ and reduced it to $0.05$ at 250K steps and $0$ at 400K steps. All experiments are done with Fairseq~\cite{ott2019fairseq}. 

% \begin{figure}[t]
%     \centering
%     \includegraphics[width=\linewidth]{Figures/finetune_framework.pdf}
%     \caption{The overall framework for fine-tuning}
%     \label{fig:framework.finetune}
% \end{figure}
\begin{table*}[t]
\begin{center}
\small
\scalebox{0.89}{
\begin{tabular}{rcccccccccccc}
\toprule

\bf Languages& 
\mf{En-Gu} &
\mf{En-Kk} &
\mf{En-Vi} &
\mf{En-Tr} & 
\mf{En-Ja} &
\mf{En-Ko}  
\\

\bf Data Source & 
\mf{WMT19}   & %gu
\mf{WMT19}   & %kk
\mf{IWSLT15} & %vi
\mf{WMT17}   & %tr
\mf{IWSLT17} & %ja
\mf{IWSLT17}  %ko
\\

\bf{Size} &
\smf{10K}  & %gu
\smf{91K}  & %kk
\smf{133K} & %vi
\smf{207K} & %tr
\smf{223K} & %ja
\smf{230K}   %ko
\\

\bf Direction & 
$\leftarrow$ & $\rightarrow$ &
$\leftarrow$ & $\rightarrow$ &
$\leftarrow$ & $\rightarrow$ &
$\leftarrow$ & $\rightarrow$ &
$\leftarrow$ & $\rightarrow$ &
$\leftarrow$ & $\rightarrow$ \\

\midrule
\bf Random &
0.0 & 0.0 & %gu
0.8 & 0.2 & %kk
23.6 & 24.8 & %vi
12.2 & 9.5 & %tr
10.4 & 12.3 & %ja
15.3 & 16.3   %ko
\\

\bf mBART25 &
\bf 0.3 & \bf 0.1 & %gu
 \bf 7.4 & \bf 2.5 & %kk
\bf 36.1 & \bf 35.4 & %vi
\bf 22.5 &\bf 17.8 & %tr
\bf 19.1 & \bf 19.4 & %ja
\bf 24.6 & \bf 22.6   %ko
\\
\midrule
\midrule

\bf Languages&
\mf{En-Nl} &
\mf{En-Ar} &
\mf{En-It} &
\mf{En-My} &
\mf{En-Ne} &
\mf{En-Ro}  
\\

\bf Data Source & 
\mf{IWSLT17} &  %nl
\mf{IWSLT17} & %ar
\mf{IWSLT17} & %it
\mf{WAT19}   & %my
\mf{FLoRes}  & %ne
\mf{WMT16}     %ro
\\

\bf{Size} &
\smf{237K} & %nl
\smf{250K} & %ar
\smf{250K} & %it
\smf{259K} & %my
\smf{564K} & %ne
\smf{608K}   %ro
\\

\bf Direction & 
$\leftarrow$ & $\rightarrow$ &
$\leftarrow$ & $\rightarrow$ &
$\leftarrow$ & $\rightarrow$ &
$\leftarrow$ & $\rightarrow$ &
$\leftarrow$ & $\rightarrow$ &
$\leftarrow$ & $\rightarrow$ \\

\midrule
\bf Random &
34.6 & 29.3 & %nl
27.5 & 16.9 & %ar
31.7 & 28.0 & %it
23.3 & 34.9 & %my
7.6  & 4.3 & %ne
34.0 & 34.3  %ro
\\

\bf mBART25 &
\bf 43.3 & \bf 34.8 &  %nl
\bf 37.6 & \bf 21.6 & %ar
\bf 39.8 & \bf 34.0 & %it
\bf 28.3 & \bf 36.9 & %my
\bf 14.5 & \bf 7.4 & %ne
\bf 37.8 & \bf 37.7  %ro
\\
\midrule
\midrule

\bf Languages& 
\mf{En-Si} &
\mf{En-Hi} &
\mf{En-Et} &
\mf{En-Lt} &
\mf{En-Fi} &
\mf{En-Lv} 
\\

\bf Data Source & 
\mf{FLoRes}  & %si
\mf{ITTB}    & %hi
\mf{WMT18}   & %et
\mf{WMT19}   & %lt
\mf{WMT17}   & %fi
\mf{WMT17}     %lv
\\

\bf{Size} &
\smf{647K}  & %si
\smf{1.56M} & %hi
\smf{1.94M} & %et
\smf{2.11M} & %lt
\smf{2.66M} & %fi
\smf{4.50M}   %lv
\\

\bf Direction & 
$\leftarrow$ & $\rightarrow$ &
$\leftarrow$ & $\rightarrow$ &
$\leftarrow$ & $\rightarrow$ &
$\leftarrow$ & $\rightarrow$ &
$\leftarrow$ & $\rightarrow$ &
$\leftarrow$ & $\rightarrow$ \\

\midrule
\bf Random &
7.2 & 1.2 & %si
10.9 & 14.2 & %hi
22.6 & 17.9 & %et
18.1 & 12.1 & %lt
21.8 & 20.2 & %fi
15.6 & 12.9 % Transformer-big, dr0.3 %lv
\\

\bf mBART25 &
\bf 13.7 & \bf 3.3 & %si
\bf 23.5 & \bf 20.8 & %hi
\bf 27.8 & \bf 21.4 & %et
\bf 22.4 & \bf 15.3 & %lt
\bf 28.5 & \bf 22.4 & %fi
\bf 19.3 & \bf 15.9 %lv
\\

\bottomrule
\end{tabular}
}
\caption{\textbf{Low/Medium Resource Machine Translation} 
Pre-training consistently improves over a randomly initialized baseline, with particularly large gains on low resource language pairs (e.g. Vi-En). }% \mike{I liked having results from previous work in this table}}% In most cases, pre-training outperforms the best published numbers.
\label{tab:LowR}
\end{center}
\end{table*}

\begin{table}[t!]
\small
\begin{center}
\scalebox{0.89}{
\begin{tabular}{rcccccc}
\toprule
%\bf Data Source & \multicolumn{5}{c}{\bf WMT}\\
\bf Languages & \bf Cs & \bf Es & \bf Zh &\bf De & \bf Ru & \bf Fr\\ 
\bf Size & 11M & 15M & 25M & 28M & 29M & 41M\\
\midrule
\textsc{\bf Random} & 16.5 & 33.2 & \bf 35.0 & \bf 30.9 & \bf 31.5 & \bf 41.4 \\
\textsc{\bf mBART25} & \bf 18.0 & \bf 34.0 & 33.3 & 30.5 &  31.3 & 41.0 \\
\bottomrule

\end{tabular}}
\end{center}
\caption{\textbf{High Resource Machine Translation} where all the datasets are from their latest WMT competitions. We only evaluate our models on En-X translation. 
}%\jgu{Double check? En-Cs, I feel it is too much.}}
%En-De are trained on WMT19 and tested on WMT14
\label{tab:HighR}
\end{table}

\subsection{Pre-trained Models}
\label{sec:pt:models}
To better measure the effects of different levels of multilinguality during pre-training, we built a range of models as follows:
%models can help downstream translation tasks on a variant amount of monolingual and bi-text data; we obtain the following types of models: 
\begin{itemize}[leftmargin=*]
    \item {\bf mBART25}~
We pre-train a model on all 25 languages, using the setting described in \cref{sec:pt:learn}. 
%As noted in \cref{sec:pt:data}, 
%Even though only 25 languages are used in this pre-training model, we use the same dictionary generated by 100 languages as described in section $3.3$. 
%even though we 
%not in pretrained monolingual data to achieve better results than plain seq2seq model see section $3.2$ Zero Monolingual Data for details.
\item {\bf mBART06}~
To explore the effect of pre-training on related languages, we pretrain a model on a subset of six European languages: Ro, It, Cs, Fr, Es and En. For a fair comparison, we use $\sim1/4$ of the mBART25 batch size, which allows our model to have the same number of updates per language during pre-training.

%\mike{Why is this fair?}
\item {\bf mBART02}~~
We pre-train bilingual models, using English and one other language for four language pairs: En-De, En-Ro, En-It. We use a batch size of $\sim1/12$ of that in the mBART25.

\item {\bf BART-En/Ro}~
To help establish baseline performance levels, we also train monolingual BART models on the same En and Ro corpus only. %We use a batch size of $\sim 1/25$ of that in the mBART25.
%For example, when we train a Ro and En bilingual model, the model should have seen the same amount of Ro and En monolingual data as the mBART25 and mBART06 model. 

\item {\bf Random}~ As additional baselines, we will also include a comparison with a model randomly initialized without pre-training for each translation task. Since the sizes of different downstream datasets vary, we always grid-search the hyper-parameters (architecture, dropout, etc.) to find the best non-pretrained configuration.
\end{itemize}
All models use the same vocabulary (\cref{sec:pt:data}).
Not all tokens will frequently occur in all pre-training corpora, but later experiments show that this large vocabulary can improve generalization in multilingual settings even for unseen languages.
\section{Sentence-level Machine Translation}
\label{sec:nmt}
%In this section, we measure performance for sentence-level machine translation. 
This section shows that mBART pre-training provides consistent performance gains in low to medium resource sentence-level MT settings, including bi-text only and with back translation, and outperforms other existing pre-training schemes (\cref{sec:nmt:result}). We also present a detailed analysis to understand better which factors contribute the most to these gains (\cref{sec:nmt:analysis}), and show that pre-training can even improve performance for languages not present in the pre-training data at all (\cref{sec:no_monolingual}).

%We observe a further improvement when combined with back translation. Besides, we show our pre-training is better than other pre-trained models and provide a detailed analysis on our pre-training.  Finally, we show that our model is able to generalize well on languages do not present in the pre-training stage.
% fine-tune languages that do not present in the pre-training data. 

% demonstrate the transfer ability of the learned representations. % \mike{This needs re-writing to make more precise claims}

\subsection{Experimental Settings}
\label{sec:nmt:setting}
%We present our results separately for low/medium and high resource pairs along with setups and language resources with comparison to back translation (BT) and other popular pre-training methods. 
\paragraph{Datasets}
% As shown in Table~\ref{tab:bitext}, 
We gather $24$ pairs of publicly available parallel corpora that cover all the languages in CC25 (Table~\ref{tab:datastats}). 
%In default, we only consider En-X (X is any other 24 languages) pairs.
%amount in term of to English and resources for the 24 languages we choose.
Most pairs are from previous WMT (Gu, Kk, Tr, Ro, Et, Lt, Fi, Lv, Cs, Es, Zh, De, Ru, Fr $\leftrightarrow$ En)
%\footnote{\url{http://www.statmt.org/wmt19/}} 
and IWSLT (Vi, Ja, Ko, Nl, Ar, It $\leftrightarrow$ En)
%\footnote{\url{https://sites.google.com/site/iwsltevaluation2017/}}
competitions. We also use FLoRes pairs~\citep[][En-Ne and En-Si]{guzman-etal-2019-flores}, En-Hi from IITB~\cite{DBLP:journals/corr/abs-1710-02855}, 
%En-Ja from KFTT~\cite{neubig11kftt}, 
and En-My from WAT19~\cite{ding2018nova,ding2019towards}. We divide the datasets into three categories -- low resource ($<$1M sentence pairs), medium resource ($>$1M and $<$10M), and high resource ($>$10M).
%Figure~\ref{fig:framework.finetune}, shows how we preprocess bi-texts. 
%\input{Tables/Bitex_stats.tex}

\paragraph{Fine-tuning \& Decoding}
We fine-tune our multilingual pre-trained models on a single pair of bi-text data, feeding the source language into the encoder and decoding the target language. As shown in Figure~\ref{fig:pt:framework}, we load the pre-trained weights and train the MT model on bi-texts with teacher forcing.
For all directions, we train with $0.3$ dropout, $0.2$ label smoothing, $2500$ warm-up steps, $3\mathrm{e}{-5}$ maximum learning rate. 
% and linear learning rate decay. 
We use a maximum of $40$K training updates for all low and medium resource pairs and $100$K for high resource pairs. The final models are selected based on validation likelihood.
For decoding, we use beam-search with beam size $5$ for all directions. The final results are reported in BLEU~\cite{papineni2002bleu} with language-specific settings, see~\cref{sec:appendix}.

%\mike{Validation loss or BLEU?} \yinhan{Validation loss}
\begin{figure*}[tbph]
    \centering
    \includegraphics[width=\linewidth]{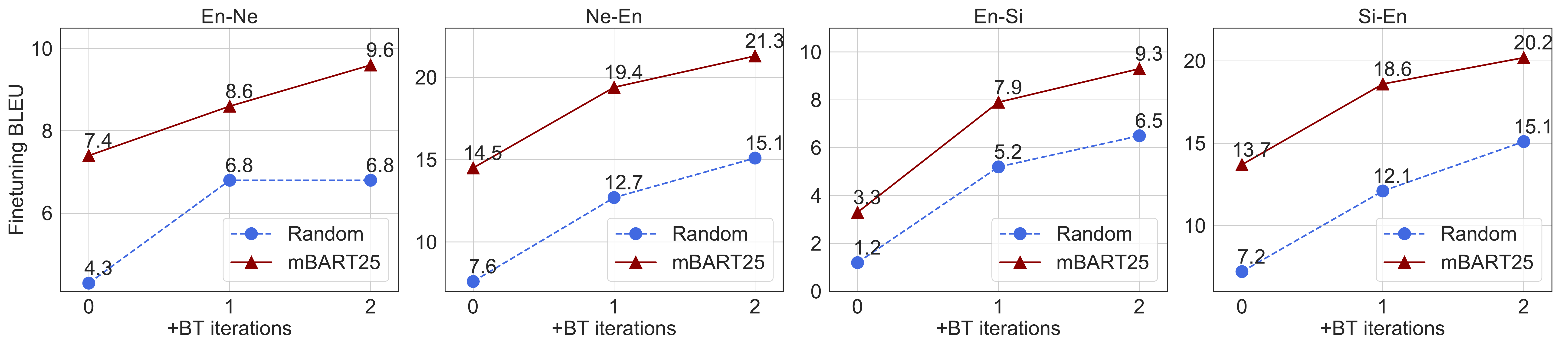}
    \caption{\textbf{Pre-training + Back Translation} on FLoRes with two iterations of BT.} %Numbers show that our gains over plain seq2seq stay the same as more iterations go
    \label{fig:FloRes}
\end{figure*}

% \begin{table*}[t]
% \begin{center}
% \begin{tabular}{lcccccc}
% \toprule
% & \bf En-Ne & {\bf En-Ne} 1it & {\bf En-Ne} 2it &\bf Ne-En &{\bf Ne-En} 1it & {\bf Ne-En} 2it\\ 
% \midrule
% \textsc{\bf Best S2S Model} & 4.3 & 6.8 & 6.8 & 7.6&12.7&15.1 \\
% \textsc{\bf Finetuned on CC25} & \bf 7.4 & \bf 8.6 & \bf9.6 & \bf 14.5& \bf 19.4& \bf 21.3 \\
% \midrule
% & \bf En-Si & {\bf En-Si} 1bt & {\bf En-Si} 2bt &\bf Si-En &{\bf Si-En} 1bt & {\bf Si-En} 2bt\\ 
% \midrule
% \textsc{\bf Best S2S Model}  & 1.2 & 5.2 & 6.5 & 7.2&12.0&15.1 \\
% \textsc{\bf Finetuned on CC25}& \bf 3.3 & \bf 7.9& \bf 9.3 & \bf 13.7 & \bf 18.6&  \bf20.2 \\
% \bottomrule

% \end{tabular}
% \end{center}
% \caption{\textbf{Pre-training + Back Translation} Plain seq2seq result VS our CC25 initialized seq2seq result on FLoRes~\cite{guzman-etal-2019-flores} with two iterations of BT. Both models use the same monolingual to generate back direction bi-text. Numbers show that our gains over plain seq2seq stay the same as more iterations go. 
% }
%\label{tab:FloRes}
%\end{table*}

\begin{table}[t]
\small
\begin{center}
\scalebox{0.9}{
\begin{tabular}{ll|ccc}
\toprule
\multicolumn{2}{c|}{\bf Pre-training} & \multicolumn{3}{c}{\bf Fine-tuning}\\
\bf Model & \bf Data & \bf En$\rightarrow$Ro & \bf Ro$\rightarrow$En & \bf +BT \\ 
\midrule
\textsc{\bf Random} & None & 34.3 & 34.0 & 36.8 \\
\midrule
\textsc{\bf XLM~\shortcite{lample2019cross}} & En Ro  & - & 35.6 & 38.5 \\
\textsc{\bf MASS~\shortcite{song2019mass}} & En Ro  & - & - & 39.1\\
\textsc{\bf BART~\shortcite{lewis2019bart}} & En  & - & - & 38.0\\
\textsc{\bf XLM-R~\shortcite{conneau2019unsupervised}} & CC100 & 35.6 & 35.8 & - \\
\midrule
\textsc{\bf BART-En} & En & 36.0 & 35.8 & 37.4 \\
\textsc{\bf BART-Ro} & Ro & 37.6 & 36.8 & 38.1 \\
\textsc{\bf mBART02}  & En Ro&\bf 38.5 & \bf 38.5 & \bf 39.9\\
\textsc{\bf mBART25}  & CC25 & 37.7 & 37.8 & 38.8 \\
\bottomrule
\end{tabular}}
\end{center}
\caption{\textbf{Comparison with Other Pre-training Approaches} on WMT16 Ro-En. 
% Our denoising seq2seq objective outperforms MASS and XLM, and bilingual pre-training on the target language pair is most effective. Backtranslation leads to additional gains in all cases, resulting in a new state-of-the for Ro-En translation.
%When compare our pretrain with other pretrain model results. Our CC25 outperform the other pretrain models, and our CCBL further push the State of The Art value on Ro-En pair.  Ro-En + BT uses monolingual data\cite{sennrich2016neural}.
% Our models outperform all other pre-trainings.
%(*) BART uses plain seq2seq generated bi-text data. 
}
\label{tab:VsPretrain}
\vspace{-10pt}
\end{table}
\subsection{Main Results}
\label{sec:nmt:result}
As shown in Table~\ref{tab:LowR}, initializing with the pre-trained mBART25 weights shows gains on all the low and medium resource pairs when compared with randomly initialized baselines. 
%Baseline model hyper-parameters are chosen using grid search over the validation sets.
We observe gains of $12+$ BLEU on low resource pairs such as En-Vi, En-Tr, and noisily aligned pairs like En-Hi.
Fine-tuning fails in extremely low-resource setting such as En-Gu, which only have roughly ~10k examples for tuning. In these settings, unsupervised translation is more appropriate, see \cref{sec:language_transfer}.

%(only 10K Wikipedia titles are available) as there lacks enough signals to kick off translation from the pre-trained model. 
%However, we show that (\cref{sec:language_transfer}) such a scenario can be nicely handled as unsupervised translation with our pre-training.

% In many cases, our model outperforms the best published approaches~\cite{chen2019facebook, Cettolo2017OverviewOT}, even though these usually involve additional techniques such as  noisy channel reranking~\cite{Yu2016TheNN}, re-ranking~\cite{shen2004discriminative}, additional parallel data and model ensembling. % \mike{and backtranslation?} \yinhan{usually we cant beat back-translation directly, we are tied to them but gain can accumulate when we add them on}

For high resource cases (Table~\ref{tab:HighR}), we do not observe consistent gains, and pre-training slightly hurts performance when $>$25M parallel sentence are available. When a significant amount of bi-text data is given, we suspect that supervised training {\bf washes out} the pre-trained weights completely. 
%with more data is able to learn the representations which pre-training learns.

\paragraph{+ Back Translation}

Back-translation~\citep[BT,][]{sennrich-etal-2016-improving} is a standard approach to augment bi-text with target side monolingual data. 
We combine our pre-training with BT and test it on low resource language pairs -- En-Si and En-Ne -- using the FLoRes dataset \citep{guzman-etal-2019-flores}. %we combine pretrain with back translation and test it on FLoRes pairs.
For a fair comparison, we use the same monolingual data as \cite{guzman-etal-2019-flores} to generate BT data. 
%In~\cref{sec:no_monolingual}, we compare pre-training over back-translation. \mike{Not sure what the previous sentence means}
%which derived numbers in Table~\ref{tab:FloRes}.
Figure~\ref{fig:FloRes} shows that initializing the model with our mBART25 pre-trained parameters improves BLEU scores at each iteration of back translation, resulting in new state-of-the-art results in all four translation directions. %Besides, the gaps over plain Seq2Seq baseline persist, as more iterations of back-translation added. 

\paragraph{v.s. Other Pre-training Approaches}
We also compare our pre-trained models with recent self-supervised pre-training methods, as shown in Table~\ref{tab:VsPretrain}. 
We consider En-Ro translation, the only pair with established results.
Our mBART model outperforms all the other pre-trained models, both with and without BT augmentation. 
We also show comparisons with the conventional BART model trained on the same En and Ro data only. Both have improvements over baselines, while worse than mBART results, indicating pre-training in a multilingual setting is essential.  
%MASS~\cite{song2019mass} in this table is trained on our reproduced code on the same En-Ro monolingual data. 
%We controlled batch size the same for our mBART02~(En-Ro) and MASS pre-training. 
Moreover, combining BT leads to additional gains, resulting in a new state-of-the-art for Ro-En translation.% based on our mBART02 model.%\mike{Needs lots of clarification on what the differences are between the pre-training settings, and why we are not using MASS's published numbers} \yinhan{mass used to have a number on their git and paper, but they deleted the number later... so there is no official result on supervised mt from mass}

\subsection{Analysis}
\label{sec:nmt:analysis}

We also present additional analysis, to better quantify when our pre-training helps. %We control the monolingual data to have a fair comparison with back translation. 

\paragraph{How many languages should you pre-train on?}
We investigate when it is helpful for pre-training to include languages other than the targeted language pair that will be used during fine tuning. Table~\ref{tab:bilingualVScc25} shows performance on four X-En pairs.
Pre-training on more languages helps most when the target language monolingual data is limited (e.g. En-My, the size of My is around $0.5\%$ of En).
%, and mBART25 outperform the mBART02 model.
%\mike{Should this be source side, since English is always the target?} \mike{Do we have results for En-X?} \yinhan{all the results in table 6 are En-X direction} \mike{That isn't what the caption says...}. \yinhan{correct the caption of table}

In contrast, when monolingual data is plentiful (De, Ro), pre-training on multiple languages slightly hurts the final results (<$1$ BLEU). %It is understandable as multilingual also bring the noise to the model. 
In these cases, additional languages may reduce the capacity available for each test language.
Additionally, the fact that mBART06 performs similar to mBART02 on Ro-En suggests that pre-training with similar languages is particularly helpful.

\begin{figure}[t]
    \centering
    \includegraphics[width=0.95\linewidth]{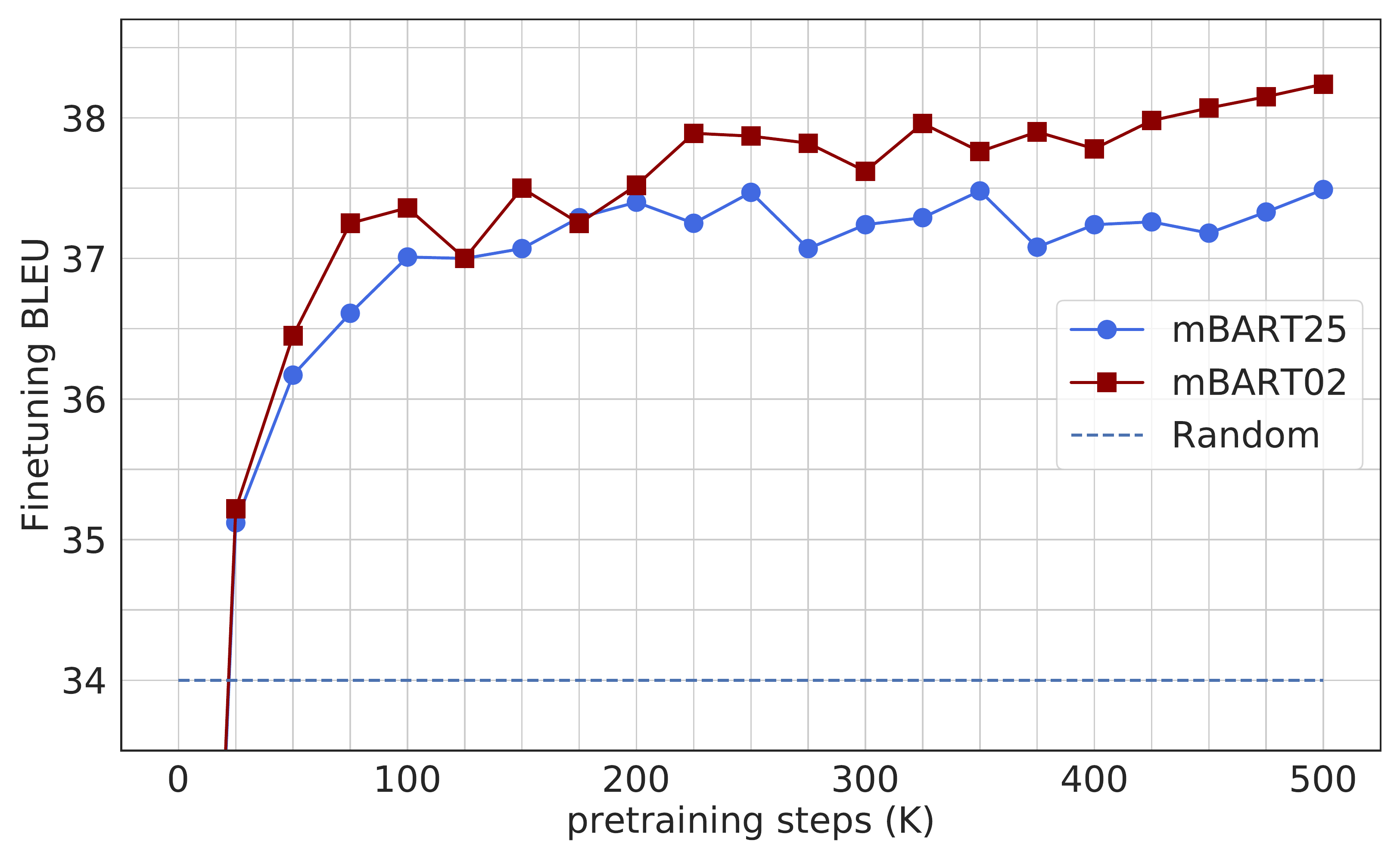}
    \caption{\textbf{Fine-tuning curves for Ro-En along with Pre-training steps}. Both mBART25 and mBART02 outperform the best baseline system after 25K steps. }% We omit the score without pre-training ($\sim 25.5$) for better visualization. }%\mike{What's the 'random' line on the graph, if it's not the 'score without pre-training'?}
    \label{fig:trend:roen}
\end{figure}

\paragraph{How many pre-training steps are needed?}
We plot Ro-En BLEU score v.s. Pre-training steps in Figure~\ref{fig:trend:roen}, where we take the saved checkpoints (every 25K steps) and apply the same fine-tuning process described in \cref{sec:nmt:setting}. Without any pre-training, our model overfits and performs much worse than the baseline. However, after just 25K steps (5\% of training), both models outperform the best baseline. The models keep improving by over $3$ BLEU for the rest of steps and have not fully converged after 500K steps. mBART25 is consistently slightly worse than mBART02.
% We can see that mBART25 stops improving since 200K steps. However, both mBART06 and RoEn bilingual still have potentials to improve as shown in Figure~\ref{fig:trend:roen}. We also observe that mBART06 and En-Ro bilingual has similar performance which is better than mBART25 since the 1st epoch.  
\begin{table}[t]
\small
\begin{center}
\begin{tabular}{lcccc|c}
\toprule
% \multirow{2}{*}{\bf Model} & \bf De & \bf Ro  & \bf It & \bf My \\ 
% &\multicolumn{4}{c}{\emph{X-En BLEU }}\\
% \midrule
% \textsc{\bf mBART02} & \bf 31.3 & \bf 38.5 & 39.7 & 36.5 \\
% \textsc{\bf mBART06} & \bf - & \bf 38.5 & 39.3 & - \\
% \textsc{\bf mBART25} & 30.5 & 37.7 & \bf 39.8 & \bf 36.9\\
% \midrule
% \multirow{2}{*}{\bf Mono Data} &\multicolumn{4}{c}{\emph{GiB}}\\
% & 66.6 & 61.4 & 30.2 & 1.6\\
% \bf Language & \bf Size/GB & \bf mBART02 & \bf mBART06 & \bf mBART25 \\
% \midrule
% \bf De & 66.6 & \bf 31.3 & - & 30.5 \\
% \bf Ro & 61.4 & \bf 38.5 & \bf 38.5 & 37.7\\
% \bf It & 30.2 & 39.7 & 39.3 & \bf 39.8\\
% \bf My & 1.6 & 36.5 & - & \bf 36.9\\
% \midrule
% \bf En & 300.8 & - & - & - \\
% \bottomrule
\bf Languages & \bf De & \bf Ro & \bf It & \bf My & \bf En \\
\midrule    
\bf Size/GB  & 66.6 & 61.4 & 30.2 & 1.6 & 300.8\\
\midrule
\bf mBART02 & \bf 31.3 & \bf 38.5 & 39.7 &  36.5\\
\bf mBART06 & - & \bf 38.5 & 39.3 & - \\
\bf mBART25 & 30.5 & 37.7 & \bf 39.8 & \bf 36.9\\
\bottomrule
\end{tabular}
\end{center}

\caption{\textbf{Pretraining Languages} on En-X translation. 
% \mike{Can we add in a no pretraining column?}
% \mike{Can we also add a column for the amount of bitext?}
The size refers to the size of monolingual data for X. The size of En is shown as reference. All the pretrained models were controlled to see the same number of English instances during training.} %The numbers show a trend that when there is a significant amount of monolingual data, bilingual pretrain is better; when monolingul data is limited, a multilingual pretrain is better.}
\label{tab:bilingualVScc25}
\end{table}
\begin{table}[t]
\small
\begin{center}
%\scalebox{0.80}{
\begin{tabular}{lcccc}
\toprule
\bf \multirow{2}{*}{Models} & \multicolumn{2}{c}{\bf En-My} & \bf Training Cost\\
& $\leftarrow$ & $\rightarrow$ & \bf GPU hours \\
\midrule
\bf Random~\shortcite{chen2019facebook} & 23.3 & 34.9 & 5\\ 
\bf \ \ \ \ + BT &  32.0  & 37.7 & 5 + 300 + 350 \\
\bf mBART02  & 29.1 & 37.8 & 300$\sim$3000 + 40  \\
\bf \ \ \ \ + BT & \bf 34.9  & \bf 39.2 & -\\
\bottomrule
\end{tabular}%}
\end{center}
\caption{Comparison with Back-Translation on My-En translation using same mono-lingual data. We also estimate the computational costs for both pre-training and back-translation based on Nvidia V100 GPUs.
}
\label{tab:bt_comparison}
\end{table}

\begin{table*}[t]
\small
\begin{center}
\scalebox{0.95}{
% <<<<<<< overleaf-2020-01-10-0531
% \begin{tabular}{ccccccc}
% \toprule
% & \multicolumn{6}{c}{\emph{Fine-tuning Data}}\\
% \emph{Pretraining Data} & \bf Nl-En & \bf En-Nl & \bf Ar-En & \bf En-Ar & \bf Nl-De & \bf De-Nl\\ 
% \midrule
% \textsc{\bf -}    & 34.6 & 29.3 &27.5 &16.9&21.3&20.9 \\
% \textsc{\bf En, Ro} & 41.4& 34.5 & 34.9&21.2&26.1&25.4\\
% \textsc{\bf mBART06: En, Ro, It, Cs, Fr, Es} & 43.1 & 34.6 & 37.3& 21.1 &26.4&25.3\\
% \hline
% \textsc{\bf mBART25 (all)} & 43.3 & 34.8 & 37.6 &21.6&27.7&26.1\\
%=======
\begin{tabular}{llllllll}
\toprule
%& \multicolumn{2}{c}{\emph{Seen BPE}}&\multicolumn{2}{c}{\emph{UnSeen BPE}}&\multicolumn{2}{c}{\emph{No Mono Pair}}\\ 
& \bf Monolingual &\bf Nl-En & \bf En-Nl & \bf Ar-En & \bf En-Ar & \bf Nl-De & \bf De-Nl\\ 
\midrule
\textsc{\bf Random} 
&None
&34.6 (-8.7)   
&29.3 (-5.5)     
&27.5 (-10.1)
&16.9 (-4.7)   
&21.3 (-6.4)
&20.9 (-5.2)\\
\midrule
\textsc{\bf mBART02}
&En Ro
& 41.4 (-2.9)     
& 34.5 (-0.3)    
& 34.9 (-2.7) 
& 21.2 (-0.4)   
& 26.1 (-1.6)
& 25.4 (-0.7)\\
\textsc{\bf mBART06} 
& En Ro Cs It Fr Es
& 43.1 (-0.2)    
& 34.6 (-0.2)    
& 37.3 (-0.3)
& 21.1 (-0.5)   
& 26.4 (-1.3)
& 25.3 (-0.8)\\
\textsc{\bf mBART25} 
& All
& \bf 43.3 & \bf 34.8 & \bf 37.6 & \bf 21.6& \bf 27.7&\bf 26.1\\

\bottomrule
\end{tabular}}
%\caption{\textbf{Generalize to Unseen Languages} fine-tuning on language-pairs without pre-training on them. \jgu{Can we compute how many BPE tokens are shared @Yinhan}}

\end{center}
% <<<<<<< overleaf-2020-01-10-0531
\caption{\textbf{Generalization to Unseen Languages} 
Language transfer results, fine-tuning on language-pairs without pre-training on them.
mBART25 uses all languages during pre-training, while other settings contain at least one unseen language pair. For each model, we also show the gap to mBART25 results.
%We find large gains from pre-training on English-Romanian, even when translating a distantly related unseen language (Arabic) and two unseen languages (German and Dutch). The best results are achieved when pre-training includes both test languages, however pre-training on other languages is surprisingly competitive.
}
% %For Nl-En and Ar-En there is no monolingual data in mBART06 or EnRo bilingual pretrain model for Nl/Ar side, but we can still see the gain in BLEU when compare it to plain S2S. For De-Nl, even though there is no monolingual data for either side, pretrained model still outperform plain S2S by ~5 BLEU. 

\label{tab:nl-en-ar}
\end{table*}
\begin{figure}[t]
    \centering
    \includegraphics[width=0.95\linewidth]{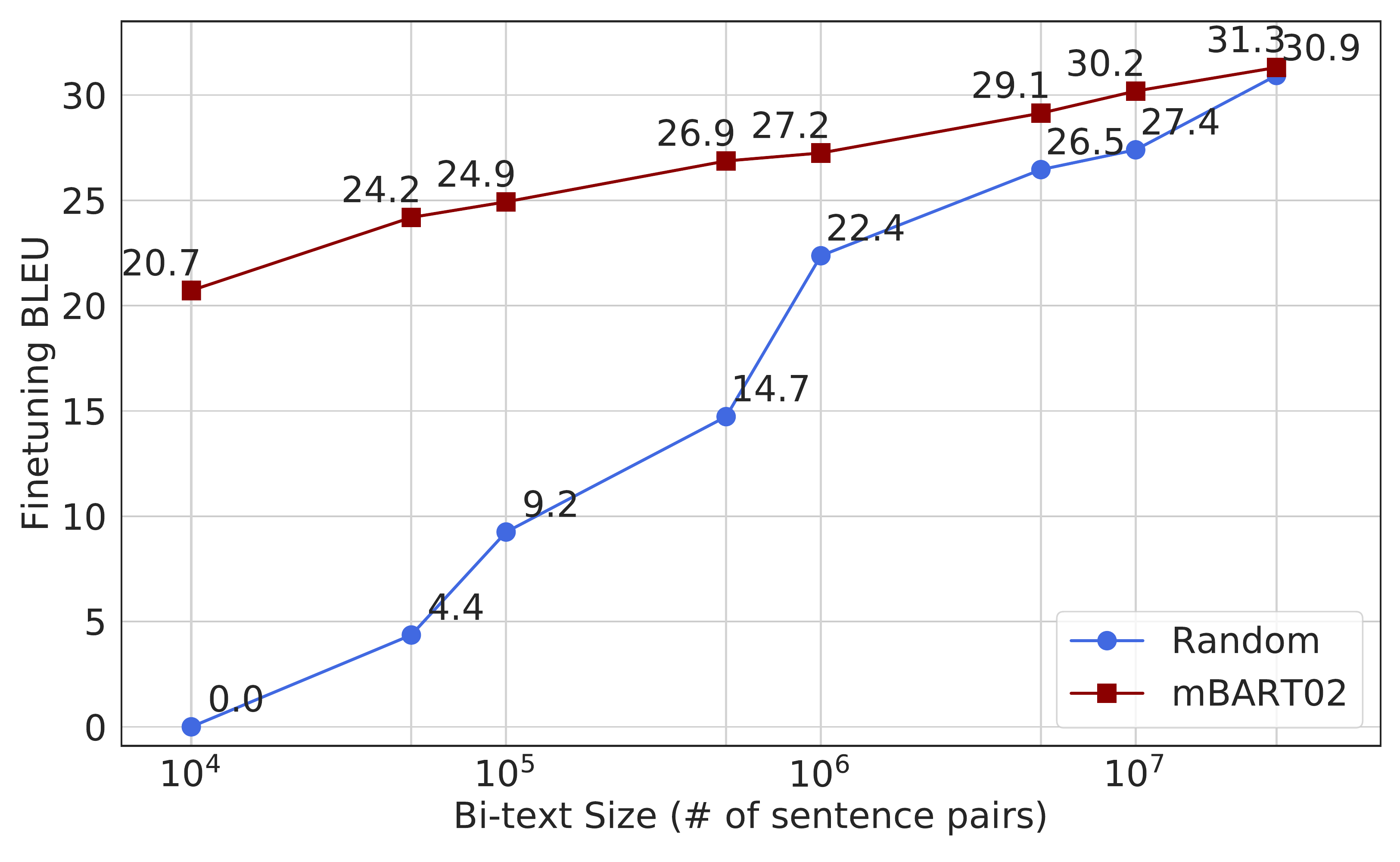}
    \caption{\textbf{Fine-tuning curves for En-De along with size of bitext.} The x-axis is on a log scale.}
    \label{fig:trend:ende}
\end{figure}

\paragraph{How does the size of bitexts inference the gain from pre-training?} 
Tables~\ref{tab:LowR} and~\ref{tab:HighR} show that pre-training consistently improves for low and medium resource language pairs. 
To verify this trend, we plot performance for differing sized subsets of the En-De dataset.
%, to control for variations between languages and domains. 
More precisely, we take the full En-De corpus ($28$M pairs) and randomly sample 10K, 50K, 100K, 500K, 1M, 5M, 10M datasets. We compare performance without pre-training to the mBART02 results, as shown in Figure~\ref{fig:trend:ende}.
%Then, we train the MT models w/o the pre-training where we use the bilingual model (mBART02 in Table~\ref{tab:bilingualVScc25}).
%For random initialized models, we select the best architecture for each size with grid search; We use the same architecture as pre-training for the mBART02 models. We plot our results in Figure~\ref{fig:trend:ende}. 
The pre-trained model is able to achieve over 20 BLEU with only 10K training examples, while the baseline system scores 0. Unsurprisingly, increasing the size of bi-text corpus improves both models. Our pre-trained model consistently outperforms the baseline models, but the gap reduces with increasing amounts of bi-text, especially after 10M sentence pairs. This result confirms our observation in \cref{sec:nmt:result} that our pre-training does not help translation in high-resource pairs. 

%We can see when parallel bi-text is under 3M pairs, our pre-training model initialized seq2seq can improve the BLEU score. 
%Even though the pre-training is done on monolingual data, the transfer learning still helps for bilingual data downstream tasks. However, for high resource language pair, the pre-trained parameters are washed out in the parallel data fine-tuning stage. 
%We are here to investigate up to much bi-text data, pre-training can help. We use En-De pair for experiments, 

\paragraph{Is pre-training complementary to BT?~~}
%\mike{Given that you can use them both together, I would say something like "complementary to" instead of "better than"}
Figure~\ref{fig:FloRes} presents that our pre-trained models can be combined with iterative back-translation (BT) on additional data, however, it is still not a fair comparison.
Table~\ref{tab:bt_comparison} shows the results when using {\bf same} monolingual data where we use $79$M En and $29$M My sentences following ~\newcite{chen2019facebook}. 

With the same amount of monolingual corpus, mBART pre-training achieves the same performance on En$\rightarrow$My as BT, while still $3$ BLEU worse on My$\rightarrow$En. We suspect BT benefits from bigger monolingual data (En). 
%achieves almost the same gains as BT with the same network architecture. 
Moreover, 
combining mBART02 model with BT, we see further gains even with same monolingual data. 
Besides, we also provide estimated training costs where BT has a longer pipeline involving training a baseline system (5h), translating monolingual data (300h) and formal training (350h). Instead, most of training costs of mBART lies in the pre-training part and can be easily adjusted to be more efficient.
%Here we provide a fair comparison study on Bilingual Pretraining VS Back Translation. We use the same monolingual data My as that in WAT19 En-My pair. 
%\jgu{TODO}

\subsection{Generalization to Languages NOT in Pre-training}
%\subsection{Generalization to Languages without Pre-training}
\label{sec:no_monolingual}

%A counter-intuitive phenomena w
%In this section, we further investigate the ability of our model to transfer to languages not present in the pre-training stage. 
%\mike{I would move this section to before the Analysis section}
In this section, we show that mBART can improve performance even with fine tuning for languages that did not appear in the pre-training corpora, suggesting that the pre-training has language universal aspects, especially within the parameters learned at the Transformer layers.

%We observe improvements, even on languages that are distantly related to those used in the pre-training.
%We observe that a pre-trained model is able to fine-tune and perform well on languages that are never seen\footnote{Due to the shard vocabulary, BPEs token embedding from unseen languages are randomly initialized.} during pre-training stage. even on distantly related language pairs. 

\paragraph{Experimental Settings}
We analyze the results of three pairs: Nl-En, Ar-En and De-Nl %(the only non-English translation task in this work) 
using the pre-trained mBART25, mBART06 and mBART02 (EnRo) models. 
During pre-training, mBART06 and EnRo Bilingual do not contain Arabic (Ar), German (De) or Dutch (Nl) data, but all languages are in mBART25. 
%Monolingual data of Ar, De, Nl are never fed to mBART02 and mBART06 models, but all of three have been learned in mBART25. 
Both De and Nl are European languages and are related to En, Ro and other the languages in mBART06 pre-training data. % In contrast, Ar is distantly related to En or other European languages. % and does not share even characters with these languages. 

\paragraph{Results}
%First, not surprisingly, fine-tuning over mBART25 achieves the best performance for all pairs with significant gains over the best plain models. However, the gap is extremely marginal between pre-trained models, especially for mBART06
%($\Delta\approx0.2\sim1$ BLEU). 

mBART25 uses all languages during pre-training, but other settings contain at least one unseen language. We find large gains from pre-training on English-Romanian, even when translating a distantly related unseen language (Arabic) and two unseen languages (German and Dutch). The best results are achieved when pre-training includes both test languages, however pre-training on other languages is surprisingly competitive. 

\paragraph{Unseen Vocabularies}
Arabic is distantly related to the languages in mBART02 and mBART06, and its use of a disjoint character set means that it word embeddings will be largely untrained.
However, we obtain similar improvements on Ar-En pairs to those on Nl-En. This result suggests that the pre-trained Transformer layers  learn universal properties of language that generalize well even with minimal lexical overlap.

\paragraph{Unseen Source or Target Languages}
Table~\ref{tab:nl-en-ar} shows different performance when the unseen languages are on the source side, target side, or both sides. If both sides are unseen, the performance (in terms of difference from mBART25) is worse than where at least one language is seen during pre-training. Furthermore, although the En-X pairs perform similarly, mBART06 outperforms mBART02 by a margin on X-En pairs. Fine-tuning unseen languages on source side is more difficult, deserving more extensive future study. 

\section{Document-level Machine Translation}
\label{sec:dnmt}
We evaluate mBART on document-level machine translation tasks, where the goal is to translate segments of text that contain more than one sentence (up to an entire document).
During pre-training, we use document fragments of up to 512 tokens, allowing the models to learn dependencies between sentences. We show that this pre-training significantly improves document-level translation.

%Since all of our pretrained models are trained on multiple consecutive sentences, which is that sentences are packed as many as possible in each instance before either reaching the maximum length (512) or hitting the document boundary. Therefore, it is quite natural to fine-tune on document-level machine translation task.
%After pre-training, our model outperforms the state-of-the-art on Zh-En TED15 and beats our own pre-trained sent-level model on En-De WMT19. 

\subsection{Experimental Settings}
\label{sec:dnmt:setting}
\begin{table}[t]
\small
\begin{center}
\begin{tabular}{lrrr}
\toprule
\bf Datasets & \bf \# Docs & \bf \# Insts & \bf \# Sents  \\
\midrule
\bf WMT19 En-De & 77K & 171K & 3.7M  \\
\bf TED15 Zh-En & 1.7K& 6.5K & 0.2M\\
\bottomrule
\end{tabular}
\end{center}
\caption{\textbf{Statistics for the Document-level Corpus} of WMT19 En-De and TED15 Zh-En. \# of instances is the \# of training examples in document model.} %\mike{Why isn't the number of training examples the same as the number of documents?}}
\vspace{-10pt}
\label{tab:DocMT-statics}
\end{table}
\begin{table*}[t]
\small
\begin{center}
% \begin{tabular}{lccc}
% \toprule
% \multirow{3}{*}{\bf Model}&\multicolumn{3}{c}{\bf{En-De WMT19}}\\

% & \bf \emph{Sent Model} & \bf \emph{Doc Model} & \bf {Pairs} \\

% &\multicolumn{2}{c}{\emph{SentEval / DocEval}}\\
% \midrule
% \textsc{\bf S2S} & 34.3/35.9 & -/7.7* & \multirow{2}{*}{3.7M} \\%&- & 22.0 &  \bf{29.9}\\
% \textsc{\bf mBART25 } & 36.4/38.0 &  \bf{37.1/38.5}\\%&24.0& 3.2 &  29.6\\
% \midrule
% \multirow{3}{*}{\bf Model}&\multicolumn{3}{c}{\bf{Zh-En TED15}}\\
% & \bf \emph{Sent Model} & \bf \emph{Doc Model} & \bf {Pairs}\\
% &\multicolumn{2}{c}{\emph{DocEval}}\\
% \midrule
% \textsc{\bf S2S} & 22.0& 3.2  & \multirow{3}{*}{211K}\\
% \textsc{\bf HAN }&- &24.0\\
% \textsc{\bf mBART25 }&\bf{29.9}&  29.6 \\
% \bottomrule
% \end{tabular}
\scalebox{0.95}{
\subtable[Sentence- and Document-level BLEU scores on {\bf En-De}]{
    \begin{tabular}{rcccc}
        \toprule
         \multirow{2}{*}{\bf Model}& \multicolumn{2}{c}{\bf Random} & \multicolumn{2}{c}{\bf mBART25} \\
         & s-BLEU & d-BLEU & s-BLEU & d-BLEU \\  
         \midrule
       \bf Sent-MT & 34.5 & 35.9 & 36.4 & 38.0\\
       \bf Doc-MT  & $\times$ & 7.7 & \bf 37.1 & \bf 38.5\\
        % & \bf Random & \bf mBART25 & \bf Random & \bf mBART25 \\
        % \midrule
        % BLEU (Sent) & 34.3 & 36.4 & $\times$ & \bf 37.1 \\
        % BLEU (Doc)  & 35.9 & 38.0 & 7.7 & \bf 38.5 \\
        \bottomrule
    \end{tabular}
}}
\scalebox{0.95}{
\subtable[Document-level BLEU scores on {\bf Zh-En}]{
    \begin{tabular}{rcc|c}
        \toprule
        \bf \multirow{2}{*}{Model} & \bf Random  & \bf mBART25 & \bf HAN~\shortcite{miculicich-etal-2018-document}\\
        & d-BLEU & d-BLEU & d-BLEU \\
        \midrule
        \bf Sent-MT & 22.0 & 28.4 & - \\
        \bf Doc-MT & 3.2 & \bf 29.6 & 24.0 \\
        % & \multicolumn{2}{c}{\bf Sent-MT} & \multicolumn{2}{c}{\bf Doc-MT} & \multirow{2}{*}{\bf HAN} \\
        % & \bf Random & \bf mBART25 & \bf Random & \bf mBART25 \\
        % \midrule
        % BLEU  & 22.0 & \bf 29.9 & 3.2 & 29.6 & 24.0 \\
        \bottomrule
    \end{tabular}
}
}
\end{center}
\caption{\textbf{Document-Level Machine Translation} on En-De and Zh-En.
%Our document-level pre-training gives large gains over previous  work on translating documents, outperforming seq2seq models and the specialized HAN architecture \cite{miculicich-etal-2018-document}.}
%En-De models were trained on wmt19 document data only and tested on wmt19. We provide document-level evaluation and sentence-level evaluation for En-De and document evaluation only for Zh-En and compared with results from Hierarchical Attention Networks~\cite{miculicich-etal-2018-document}. 
($\times$) The randomly initialized Doc-MT model cannot produce translations aligned to the original sentences, so only document evaluation is possible.}% \mike{Why are the captions above the table?}}

%where we merged its sentence outputs into documents and evaluated them with BLEU. }
\label{tab:DocMT-result}
\end{table*}
\paragraph{Datasets}
We evaluate performance on two common document-level MT datasets: WMT19 En-De and TED15 Zh-En (statistics in Table~\ref{tab:DocMT-statics}). For En-De, we use the document data from WMT19 to train our model, without any additional sentence-level data; Zh-En dataset is from the IWSLT 2014 and 2015 evaluation campaigns~\cite{cettolo2012wit3,cettolo2015iwslt}. Following~\newcite{miculicich-etal-2018-document}, we use 2010-2013 TED as the test set. % Table~\ref{tab:DocMT-statics} presents statistics for our training sets.
% \yinhan{TODO: Add data stats here} 
 
 \paragraph{Pre-processing}
We use the same pre-processing as that in pre-training. 
 %We pack as many sentences as possible for each instance and stop when either source or target language reaches 512 max sequence length or meets document boundary. 
 For each block, sentences are separated by end of sentence symbols (</S>) and the entire instance is ended with the specific language id (<LID>). The numbers of segmented instances are also shown in Table~\ref{tab:DocMT-statics} where on average, every document is split into 2-4 instances. 
 
 \paragraph{Fine-tuning \& Decoding}
We use the same fine-tuning scheme as for sentence-level translation (\cref{sec:nmt:setting}), % \mike{is this true?}. 
 %Our initial experiments find that pre-training allows us to avoid 
 without using any task-specific techniques developed by previous work~\cite{miculicich-etal-2018-document,li2019pretrained}, such as constrained contexts or restricted attention.
 For decoding, we simply pack the source sentences into blocks, and translate each instance block autoregressively. The model does not know how many sentences to generate in advance and decoding stops when <LID> is predicted. We use beam size 5 by default.

\paragraph{Baselines \& Evaluation}
We train 4 models: a document-level (Doc-) MT model (\cref{sec:dnmt:setting}) and a corresponded sentence-level (Sent-) MT model (\cref{sec:nmt:setting}) as the baseline, both with and without pre-training. We use mBART25 as the common pre-trained model for En-De and Zh-En.
For En-De, even though our mBART25 Doc-MT model decodes multiple sentences together, the translated sentences can be aligned to the source sentences, which allows us to evaluate BLEU scores both on sentence-level (s-BLEU) and document-level (d-BLEU)~\footnote{Standard BLEU scores match n-grams at sentence-level. We also consider document-level where we match n-grams over the whole document resulting in a slightly higher score.}. For Zh-En, however, we cannot produce the same number of translated sentences as the reference 
%align the translated sentences to the source 
due to alignment errors in the test data. % \luke{What does this mean?} 
We only provide the d-BLEU scores on this direction.

%in Table~\ref{tab:DocMT-de} second block. 
We also compare our models with Hierarchical Attention  Networks~\citep[HAN,][]{miculicich-etal-2018-document} on Zh-En, which is the state-of-the-art non-pretraining approach for document-level translation for this pair. They combine two layers of attention -- first within and then across sentences. 

\subsection{Main Results}
We show the main results for both En-De and Zh-En are presented in Table~\ref{tab:DocMT-result}.
\paragraph{Random v.s. Pre-trained}
The MT models initialized with pre-trained weights outperform randomly initialized models by large margins, for both sentence-level and document-level training. 
Our mBART25 models (both Sent-MT and Doc-MT) also outperform HAN~\cite{miculicich-etal-2018-document}\footnote{d-BLEU is recomputed from the provided system output.}, despite the fact that they are not customized for document-level MT in any way.
\paragraph{Sent-MT v.s. Doc-MT}~ For cases (En-De, En-Zh), the mBART25 Doc-MT models outperform themselves fine-tuned at sentence-level by a margin, which is completely opposite for models without pre-training. 
For both datasets, randomly initialized Doc-MT fail to work, resulting in much worse results than the sentence-level models. Such large performance gaps indicate that pre-training is \textit{critical} for document level performance. It is in general difficult to collect high quality document-level data in large quantities, suggesting that pre-training may be a strong strategy for future work.  We also include a sampled example in \cref{sec:example}.

%As one potential reason, highdocument-level MT has been suffering from the lack of high quality parallel data. It is very likely to overfit on spurious correlations/dependencies existed in the training pairs. Pre-training, as a strong prior, has helped build correct attentions at document level before fine-tuning.

% \paragraph{Analysis}
% Document level machine translation has been suffering from the lack of high quality parallel data. Results on both datasets suggest that during pre-training, a large amount of monolingual data trained on document level can help the model learn long range contexts, and these learned transformer attention weights from monolingual data can be transferred to bi-text datasets across languages. \jgu{TODO: More analysis or explanation}

\begin{figure*}[t]
    \centering
    \includegraphics[width=0.95\linewidth]{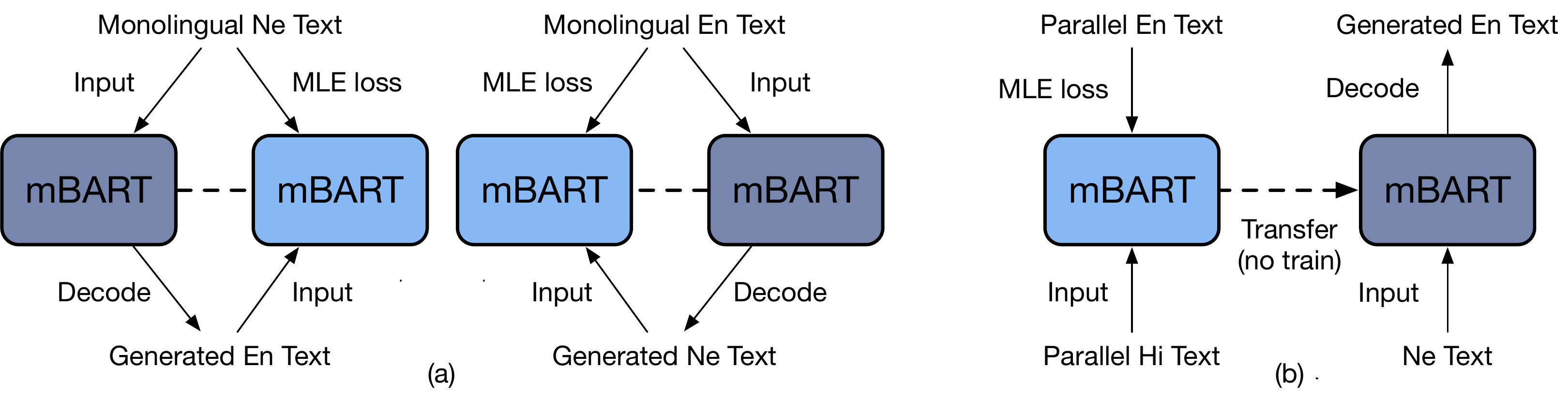}
    \caption{Illustrated frameworks for unsupervised machine translation via (a) back-translation (b) language transfer where Ne-En is used as an example. For both cases, we initialize from multilingual pre-training (e.g. mBART25).} %\mike{I can't understand this figure}}
    \label{fig:unsupmt_framework}
\end{figure*}

\section{Unsupervised Machine Translation}
\label{sec:unsup}
In addition to supervised machine translation, we also evaluate our model on tasks where no bi-text is available for the target language pair. % (which we call unsupervised). 
We define three types of \textit{unsupervised} translation:
%We consider two kinds of unsupervised translation. 
\begin{enumerate}[leftmargin=*]
    \item No bi-text of any kind is given. A common solution is to learn from back-translation (BT) ~\citep{artetxe2017unsupervised,lample2018phrase}. We show that mBART provides a simple and effective initialize scheme for these methods. 

  \item No bi-text for the target pair is available, but the target languages both appear in bi-text corpora for other language pairs. Previous work has shown that zero-shot transfer is possible via massively multi-lingual MT ~\citep{johnson2017google,gu2019improved} or distillation through pivoting~\citep{chen2017teacher}. We limit our focus to building MT models for single language pairs, and leave multi-lingual pre-training for multi-lingual MT to future work.% \mike{I think this bullet isn't clear that we're not reporting results in this setting}
  \item No bi-text for the target pair is available, but there is bi-text for translating from some other language into the target language. This is a new evaluation regime, where we will show that mBART supports effective transfer, even if the source language has no bi-text of any form.
\end{enumerate}
In this section, we demonstrate the effectiveness of multilingual pre-training in unsupervised machine translation via (1) back-translation (~\cref{sec:van-sup}) and (3) language transfer~(\cref{sec:language_transfer}). An illustration of both approaches are presented in Figure~\ref{fig:unsupmt_framework}.
 %In this situation, both X and Y monolingual corpus are used to train the pretraining model, and no X-Y bi-text is given, but Z-Y is given, where Z is a similar language to X. 

\subsection{Unsupervised Machine Translation via Back-Translation}
\label{sec:van-sup} 
%This is the most commonly studied unsupervised machine translation setting~\cite{lample2019cross,song2019mass}.
\paragraph{Datasets}
We evaluate our pre-trained models on both similar (En-De, En-Ro) and dissimilar pairs (En-Ne, En-Si), which are determined by measuring the subword units that are shared between the source and target languages. We use the same test sets as the supervised benchmarks~\cref{sec:nmt:setting}, and directly use the pre-training data (CC25) for back-translation to avoid introducing new information.

%pairs into similar and dissimilar languages. They are defined by how many percent of BPEed tokens shared between source and target languages. 

\paragraph{Learning}
Following the same procedure described in~\newcite{lample2018phrase,lample2019cross}, we first initialize the translation model with the pre-trained weights, and then learn to predict the monolingual sentences conditioned on source sentences generated by on-the-fly back-translation (BT). 
\newcite{lample2019cross} only pre-train an encoder, so perform additional de-noising training to learn a seq2seq model -- a step which is unnecessary for mBART's pre-trained seq2seq model. 
However, we do constrain mBART to only generating tokens in target language~\footnote{We mask out the output probability of predicting tokens which appear less than $1\%$ in the target monolingual corpus.} for the first $1000$ steps of on-the-fly BT, to avoid it simply copying the source text. %\mike{Details needed on here on what it means for a BPE token to be only in the target language}
%na\"ively applying the same algorithm on our model fail to produce any meaningful translations as pre-training has made the model to copy the source language to the target.
 %To recover from such local minimum, we enforce our model to kick off translation by only generating tokens in target corpus for the first 1K steps of on-the-fly BT. %Then we remove this constraint. 
\begin{table*}[t]
\begin{center}
\small
%\scalebox{0.76}{
\begin{tabular}{lcccccccc}
\toprule
\multirow{3}{*}{\bf Model} & \multicolumn{4}{c}{\bf Similar Pairs} &  \multicolumn{4}{c}{\bf Dissimilar Pairs} \\
& \multicolumn{2}{c}{\bf En-De} & \multicolumn{2}{c}{\bf En-Ro} & \multicolumn{2}{c}{\bf En-Ne} & \multicolumn{2}{c}{\bf En-Si}\\
& $\leftarrow$ & $\rightarrow$ & $\leftarrow$ & $\rightarrow$ & $\leftarrow$ & $\rightarrow$ & $\leftarrow$ & $\rightarrow$ \\
\midrule
\bf Random & 21.0 & 17.2 & 19.4 & 21.2 & 0.0 &0.0 & 0.0 & 0.0 \\
\bf XLM~\shortcite{lample2019cross} & 34.3 & 26.4 & 31.8 & 33.3 & 0.5 & 0.1 & 0.1 & 0.1\\
\bf MASS~\shortcite{song2019mass} & \bf 35.2 & 28.3 & \bf 33.1 & \bf 35.2 & - & - & - & -\\
\midrule
\bf mBART & 34.0 & \bf 29.8 & 30.5 &  35.0 & \bf 10.0 & \bf 4.4 & \bf 8.2 & \bf 3.9\\
% & \multicolumn{3}{c}{\emph{similar languages}}& \multicolumn{2}{c}{\emph{dissimilar languages}}\\
% & \bf Ro-En & \bf It-En & \bf De-En & \bf Ne-En & \bf Si-En\\ 
% \midrule
% \textsc{\bf XLM} &\bf 31.8 &  -&\bf 34.3& 0.5 & 0.1 \\
% \textsc{\bf Our Pretain} & 30.5&29.0&34.0&\bf 10.0 & \bf 8.2 \\

% & \bf En-Ro & \bf En-It & \bf En-De & \bf En-Ne & \bf En-Si\\ 
% \midrule
% \textsc{\bf XLM} &33.3&-&26.4& 0.1 & 0.1 \\
% \textsc{\bf Our Pretain} &\bf 34.1&25.8&\bf 29.8& \bf 4.4& \bf 3.9  \\

\bottomrule
\end{tabular}
%}
\end{center}
\caption{\textbf{Unsupervised MT via Back-Translation}. En-De, En-Ro are initialized by mBART02, while En-Ne, En-Si are initialized by mBART25. Our models are trained on monolingual data used in pre-training.
}
\label{tab:UnFloRes}
\end{table*}
\begin{table*}[t]
\begin{center}
\small
\scalebox{0.95}{
\begin{tabular}{rc|cccccccccccc}
\toprule
& & \multicolumn{12}{c}{\bf Fine-tuning Languages} \\
%\cmidrule{3-12}
& & 
\bf Zh & \bf Ja & \bf Ko & \bf Cs   &\bf Ro  & \bf Nl  &\bf It & \bf Ar &  \bf Hi & \bf Ne & \bf Si &\bf Gu\\ 
%\multicolumn{2}{r|}{\bf Groups} & Asian & Asian & European & European & European  & European & Arabic& Indic & Indic & Indic \\ 
%\multicolumn{2}{r|}{\bf Bitext Size} & 25M & 223K & 230K & 11M & 608K & 237K & 250K & 250K & 1.56M & 564K & 10K\\
\multicolumn{2}{r|}{\bf Domain} & News & TED & TED & News & News & TED & TED & TED & News & Wiki & Wiki & Wiki\\
\midrule
\multirow{10}{*}{\STAB{\rotatebox[origin=c]{90}{\bf Testing Languages}}}& 
\textsc{\bf Zh}  &\cgg 23.7& \cg 8.8 &\bf \cg 9.2& 2.8&  7.8 &7.0&6.8&6.2&7.2&4.2 & 5.9 & 0.0\\
& \textsc{\bf Ja} & \cg 9.9 & \cgg 19.1 & \cg \bf 12.2 & 0.9 & 4.8 & 6.4 & 5.1 & 5.6 & 4.7 & 4.2 & 6.5 & 0.0\\
& \textsc{\bf Ko} & \cg{5.8} & \cg \bf 16.9 &\cgg 24.6&5.7&8.5&  9.5 & 9.1&  8.7& 9.6 &8.8 &11.1 & 0.0\\
& \textsc{\bf Cs} & 9.3 & 15.1 &17.2&\cgg 21.6&  \bf \cg 19.5& \cg 17.0 &\cg 16.7& 16.9&13.2&15.1 & 16.4 & 0.0\\
%\textsc{\bf De} &&&20.4&&&&&25.8\\
& \textsc{\bf Ro} & 16.2 & 18.7 &17.9& \cg \bf 23.0&\cgg 37.8 & \cg 22.3& \cg 21.6&22.6&16.4&18.5&22.1& 0.0\\
& \textsc{\bf Nl} &14.4 & 30.4& 32.3&\cg 21.2&\cg 27.0& \cgg 43.3 & \cg \bf 34.1&  31.0&24.6&23.3&27.3& 0.0\\
& \textsc{\bf It} &16.9 & 25.8 & 27.8&\cg 17.1&\cg 23.4& \cg 30.2 &\cgg 39.8&  \bf 30.6&20.1&18.5&23.2& 0.0\\
& \textsc{\bf Ar} &5.8 & \bf 15.5& 12.8&12.7&12.0&  14.7 & 14.7& \cgg 37.6&11.6&13.0&16.7& 0.0\\
& \textsc{\bf Hi} &3.2 & 10.1 &9.9&5.8&6.7& 6.1 &5.0& 7.6&\cgg 23.5&\bf \cg 14.5 & \cg 13.0& \cg 0.0\\
& \textsc{\bf Ne}&2.1 & 6.7 &6.5&5.0&4.3&3.0&2.2&5.2&\cg \bf 17.9&\cgg 14.5 & \cg 10.8 & \cg 0.0\\
%\cmidrule{2-11}
& \textsc{\bf Si} & 5.0 & 5.7 & 3.8 & 3.8 & 1.3& 0.9& 0.5 & 3.5& \cg 8.1 & \cg\bf 8.9 & \cgg 13.7 &\cg 0.0\\
& \textsc{\bf Gu} & 8.2 & 8.5 & 4.7& 5.4 & 3.5 &2.1 & 0.0 &6.2&\cg \bf 13.8 & \cg 13.5 & \cg12.8 & \cgg 0.3\\
%\midrule
%& \textsc{\bf Mean} &    &    &    &    &    &     &     &     
\bottomrule
\end{tabular}
}
\end{center}
\caption{\textbf{Unsupervised MT via Language Transfer} on X-En translations. The model fine-tuned on one language pair is directly tested on another. 
%\mike{It's not obvious enough which of these is the row or column - maybe rename 'Transferring Language' as 'Test Language'} 
We use {\color{darkgray} gray} color to show the direct fine-tuning results, and {\color{gray} lightgray} color to show language transfer within similar language groups. We {\bf bold} the highest transferring score for each pair.                                                          
%We test sacrebleu on It and  Ne pair after finetuned on Cs, Ro, Ko, Nl, Ar, It-Bitext and Ne-Bitext (Sup MT)
}
\label{tab:transferlearning}
\end{table*}

\paragraph{Results}
     Table~\ref{tab:UnFloRes} shows the unsupervised translation results compared with non-pretrained models, as well as models with existing pre-training methods. Our models achieve large gains over non-pretrained models for all directions, and outperform XLM significantly for dissimilar pairs (En-Ne, En-Si) where the existing approaches completely fail. For similar pairs, our model also performs well against XLM and MASS, with the best numbers for En-X pairs.

% For similar language pairs, 
%We observe 9.5 BLEU points gain in the Ne-En, and 8.1 BLEU gain in Si-En compared with XLM. We achieve head-to-head performance for the similar language pairs (Ro-En and De-En).
 %\jgu{TODO: add details.}

\begin{table}[t]
\begin{center}
\small
\begin{tabular}{rr|rr|c}
\toprule
\bf Pairs & \bf BT & \multicolumn{2}{c|}{\bf Transfer} & \bf Combined \\

\midrule
\bf Ro$\rightarrow$En & 30.5 &  \bf Cs$\rightarrow$En &23.0 & \bf 33.9 \\
%\bf It$\rightarrow$En & 29.0 & {\bf 30.6} (Ar$\rightarrow$En) & $\rightarrow$ \\
\bf Ne$\rightarrow$En & 10.0  & \bf Hi$\rightarrow$En& 18.9 & \bf 22.1 \\
%\bf Cs$\rightarrow$En & 17.9 & {\bf 19.5} (Ro$\rightarrow$En) & $\rightarrow$ \\
\bf Zh$\rightarrow$En & 11.3  & \bf Ko$\rightarrow$En &  9.2& \bf 15.0 \\
%\bf Ko$\rightarrow$En & \bf 10.1 &  9.6 (Hi$\rightarrow$En) & $\rightarrow$ \\
%\bf Hi$\rightarrow$En & 13.8 & {\bf 14.5} (Ne$\rightarrow$En) & $\rightarrow$ \\
\bf Nl$\rightarrow$En & 28.5  & \bf It$\rightarrow$En & 34.1& \bf 35.4 \\
%\bf Ar$\rightarrow$En & \bf 17.8 & 14.7 (Nl/It$\rightarrow$En) & $\rightarrow$ \\
\bottomrule
\end{tabular}
\caption{\label{tab:transfer_vs_bt}{\bf Back-Translation v.s. Language Transfer for Unsupervised MT}. We present the best transferring scores together with the pairs transferred from.} %We also include the {\bf Combined} score, where we start from the best transferred model, and continue unsupervised translation with back-translation.}
\end{center}
\vspace{-12pt}
\end{table}
\subsection{Unsupervised Machine Translation via Language Transfer}
\label{sec:language_transfer}
The second case of unsupervised machine translation assumes the target language appears in a bi-text corpus with some other source language. 
%Since it is straightforward to obtain the reverse direction model by training over the back-translation data once we have the forward model, we only consider additional translation to the target side. 
\paragraph{Datasets}
%is slightly different from vanilla unsupervised machine translation setting. No direct bi-text data is available for a given pair but another pair, which is translated to the same target language is provided. 
We only consider X$\rightarrow$En translation, and choose the bitexts of 12 language pairs from \cref{sec:nmt:setting}, covering Indic languages (Ne, Hi, Si, Gu), European languages (Ro, It, Cs, Nl), East Asian languages (Zh, Ja, Ko) and Arabic languages (Ar). 
%Those language pairs span different domains (news, common-crawl, and TED Talks). 
\paragraph{Results}
As illustrated in Figure~\ref{fig:unsupmt_framework}~(b), we take the pre-trained mBART25 model and finetune on each language pair, and then directly apply them to the rest of pairs, as seen in Table~\ref{tab:transferlearning}. We also present the direct fine-tuning performance (\cref{sec:nmt}) on the diagonal, for reference.
We can always obtain reasonable transferring scores at all pairs over different fine-tuned models except from Gu-En where the supervised model completely fails ($0.3$ BLEU). In some cases, we can achieve similar (Cs-En) or even much better (Ne-En, Gu-En) results compared to the supervised results.

As a comparison, we also apply the same procedure on randomly initialized models without pre-training, which always ends up with $\approx\bf0$ BLEU. This indicates that multilingual pre-training is essential and produces universal representations across languages, so that once the model learns to translate one language to En, it learns to translate all languages with similar representations. We also present three examples of language transferring between Zh, Ja and Ko in \cref{sec:example}.

\paragraph{When is language transfer useful?}
Table~\ref{tab:transferlearning} also shows mixed results at each pair. First, for most pairs, language transfer works better when fine-tuning is also conducted in the same language family, especially between Indic languages (Hi, Ne, Gu).
%It is commonly easier to transfer on similar languages (e.g. Ne-En works much better on Hi-En models than other pairs).
However, significant vocabulary sharing is not required for effective transfer. For instance, Zh-En and It-En achieve the best transfer learning results on Ko-En and Ar-En, respectively. However, the vocabulary overlapping (even character overlapping) between Zh and Ko, It and Ar is low.

\paragraph{w/ Back-Translation}
We also present the comparison on 4 pairs of unsupervised MT with back-translation (BT) v.s. language transfer in Table~\ref{tab:transfer_vs_bt}. The results are also mixed.
If there exists high quality (similar languages) bi-text data, or translating between dissimilar pairs, language transfer is able to beat the conventional methods with BT. 
Furthermore, we also show promising results for combining these two techniques. In such cases, we start from the best transferred model and apply (iterative) BT on the same monolingual corpus used in pre-training. Table~\ref{tab:transfer_vs_bt} presents the results with 1 iteration of BT. For all pairs, we see improvements by combining both techniques.

%We conclude the pros and cons as follows:

%Table~\ref{tab:transferlearning} shows the transfer learning results along with vanilla unsupervised learning~\cref{sec:van-sup} and compared with supervised learning~\cref{sec:nmt} in diagonal. 
%In the settings, plain seq2seq usually results a Zero BLEU score, which fails to transfer any language knowledge. 
%We observe that for Ne-En pair, similar language pair (Hi-En) achieves a better score than supervised learning. For It-En pair, transfer learning from some related (Nl-En) and even distantly related (Ar-En) language pairs outperforms unsupervised learning. 
%We surprisingly observe that Ko-En is always good at transferring knowledge to other pairs, but hard to be transferred to. Nl-En is easily to be learned from other pairs, but hard to transfer to other distantly related language pairs. Zh-En is hard at both directions. 

\section{Related Work}
\label{sec:background}
\paragraph{Pre-training for Text Generation}
This work inherits from the recent success brought by self-supervised pre-training for NLP applications~\cite{peters2018deep,radford2018gpt,devlin2018bert,yang2019xlnet,liu2019roberta}, especially for text generation tasks~\cite{radford2019language,song2019mass,dong2019unified,raffel2019exploring,lewis2019bart} where different self-supervised objectives are designed for training 
big neural models 
on enormous unlabeled text corpora 
%\mike{Why over-parameterized? They all underfit the pre-training objectives}. 
The pre-trained models are usually used as the initialization for fine-tuning variant downstream tasks such as controllable language modeling~\cite{shirish2019ctrl}, machine translation~\cite{song2019mass}, summarization~\cite{liu2019text} and dialogue generation~\cite{zhang2019dialogpt}. In contrast to most prior work, we focus on a deep exploration of applying denoising pre-training for various translation applications.

\paragraph{Multilinguality in NLP tasks} This work is also related to the continual trend of multilingual language learning, including aligning multilingual word embeddings~\cite{DBLP:journals/corr/MikolovLS13,chen-cardie-2018-unsupervised,lample2018word} into universal space, and learning cross-lingual models~\cite{DBLP:wada,lample2019cross,conneau2019unsupervised} to exploit shared representations across languages.

For machine translation, the most relevant field is \textit{multilingual translation}~\cite{firat2016multi,viegas2016google,aharoni-etal-2019-massively,DBLP:journals/corr/abs-1907-05019} where the ultimate goal is to jointly train one translation model that translates multiple language directions at the same time, and shares representations to improve the translation performance on low-resource languages~\cite{gu-etal-2018-universal}. In this paper, we mainly focus on multilingualism in the pre-training stage and fine-tune the learned model in the standard bi-lingual scenario. Compared to multilingual translation, we do not require parallel data across multiple languages but the targeted direction, which potentially improves the scalability to low-resource languages and specific domains. Moreover, multilingual pre-training is unlikely to suffer the interference problems between dissimilar languages, which is typical for regular multilingual translation models. % \mike{Why?}

\paragraph{Document Translation} As one of the key applications, this work also links to previous efforts for incorporating document-level contexts into neural machine translation~\cite{wang-etal-2017-exploiting-cross,DBLP:journals/corr/JeanLFC17,tiedemann-scherrer-2017-neural,miculicich-etal-2018-document,doi:10.1162/tacla00029}. \newcite{li2019pretrained} is the most relevant work which also utilized pre-trained encoder (BERT) for handling longer context. However, {\bf none} of these works had shown positive results on pure Seq2Seq models at document-level, which involved task-specific techniques, and usually only worked on sentence-level translation with a constrained range of context. To the extent of our knowledge, our multilingual pre-trained model is the first-of-its-kind work that shows improved results on document-level translation with standard Seq2Seq learning. 

\paragraph{Unsupervised Translation}
This work also summarizes the previous efforts of learning to translate between languages without a direct parallel corpus, and {\bf re-defines} them as unsupervised machine translation with three categories where in this work, we only focus on applications to the first and the third kinds~(\cref{sec:unsup}). 
When no parallel corpus of any kind is available, \newcite{artetxe2017unsupervised,lample2018unsupervised,lample2018phrase} proposed to jointly learn denoising auto-encoder and back-translation from both directions, which, however, required good initialization and only worked well on similar language pairs; \newcite{wu2019extract} replaced back-translation with retrieved similar sentences from target monolingual data; \newcite{wu2019machine} solves the problem by mining sentences from Wikipedia and use them as weakly supervised translation pairs. Similar to \newcite{lample2019cross,song2019mass}, we follow the first approach and treat our pre-trained model as the initialization step.
Besides, we investigate unsupervised translation using language transfer, which is similar to \newcite{Pourdamghani2019TranslatingTA} where the authors generate translationese of the source language and train a system on high-resource languages to correct these intermediate utterances. It is also closely related to \newcite{conneau2018xnli,artetxe2019crosslingual} for cross-lingual representation learning.
%Similar in \newcite{guzman-etal-2019-flores}, we also show improved results by combining transfer with back-translation.
\section{Conclusion}
We demonstrate that multilingual de-noising pre-training is able to significantly improve both supervised and unsupervised machine translation at both the sentence level and document level. We analyze when and how pre-training is most effective and can be combined with other approaches such as back-translation. Our results also show the transfer learning ability of the learned representations from multilingual pre-training.

\label{sec:conclusion}
%\section{Future Work}
In future work, we will scale-up the current pre-training to more languages, e.g., an mBART100 model. 
The size of our model makes it expensive to deploy in production -- future work will explore pre-training more efficient models.
%Also, the current approach is still restricted to deploy in practise due to high inference costs (computation, storage and latency) which requests further research on efficient pre-training methods.

%It will also be important to study the effects of mulit-lingual fine tuning and improve performance in very low resource settings.
\label{sec:future}
\section{Acknowledgements}
\label{sec:acknowledge}
We thank Marc'Aurelio Ranzato, Guillaume Lample, Alexis Conneau, and Michael Auli for sharing their expertise on low-resource and unsupervised machine translation, Peng-Jen Chen, Jiajun Shen for details about FloRes and WAT datasets. We also thank our colleagues at FAIR and FAIAR for valuable feedback.

\bibliographystyle{acl_natbib}
\bibliography{references}

\begin{thebibliography}{57}
\expandafter\ifx\csname natexlab\endcsname\relax\def\natexlab#1{#1}\fi

\bibitem[{Aharoni et~al.(2019)Aharoni, Johnson, and
  Firat}]{aharoni-etal-2019-massively}
Roee Aharoni, Melvin Johnson, and Orhan Firat. 2019.
\newblock \href {https://doi.org/10.18653/v1/N19-1388} {Massively multilingual
  neural machine translation}.
\newblock In \emph{Proceedings of the 2019 Conference of the North {A}merican
  Chapter of the Association for Computational Linguistics: Human Language
  Technologies, Volume 1 (Long and Short Papers)}, pages 3874--3884,
  Minneapolis, Minnesota. Association for Computational Linguistics.

\bibitem[{Arivazhagan et~al.(2019)Arivazhagan, Bapna, Firat, Lepikhin, Johnson,
  Krikun, Chen, Cao, Foster, Cherry, Macherey, Chen, and
  Wu}]{DBLP:journals/corr/abs-1907-05019}
Naveen Arivazhagan, Ankur Bapna, Orhan Firat, Dmitry Lepikhin, Melvin Johnson,
  Maxim Krikun, Mia~Xu Chen, Yuan Cao, George Foster, Colin Cherry, Wolfgang
  Macherey, Zhifeng Chen, and Yonghui Wu. 2019.
\newblock \href {http://arxiv.org/abs/1907.05019} {Massively multilingual
  neural machine translation in the wild: Findings and challenges}.
\newblock \emph{CoRR}, abs/1907.05019.

\bibitem[{Artetxe et~al.(2017)Artetxe, Labaka, Agirre, and
  Cho}]{artetxe2017unsupervised}
Mikel Artetxe, Gorka Labaka, Eneko Agirre, and Kyunghyun Cho. 2017.
\newblock Unsupervised neural machine translation.
\newblock \emph{arXiv preprint arXiv:1710.11041}.

\bibitem[{Artetxe et~al.(2019)Artetxe, Ruder, and
  Yogatama}]{artetxe2019crosslingual}
Mikel Artetxe, Sebastian Ruder, and Dani Yogatama. 2019.
\newblock \href {http://arxiv.org/abs/1910.11856} {On the cross-lingual
  transferability of monolingual representations}.

\bibitem[{Cettolo et~al.(2012)Cettolo, Girardi, and Federico}]{cettolo2012wit3}
Mauro Cettolo, Christian Girardi, and Marcello Federico. 2012.
\newblock Wit3: Web inventory of transcribed and translated talks.
\newblock In \emph{Conference of European Association for Machine Translation},
  pages 261--268.

\bibitem[{Cettolo et~al.(2015)Cettolo, Jan, Sebastian, Bentivogli, Cattoni, and
  Federico}]{cettolo2015iwslt}
Mauro Cettolo, Niehues Jan, St{\"u}ker Sebastian, Luisa Bentivogli, Roldano
  Cattoni, and Marcello Federico. 2015.
\newblock The iwslt 2015 evaluation campaign.
\newblock In \emph{International Workshop on Spoken Language Translation}.

\bibitem[{Chen et~al.(2019)Chen, Shen, Le, Chaudhary, El-Kishky, Wenzek, Ott,
  and Ranzato}]{chen2019facebook}
Peng-Jen Chen, Jiajun Shen, Matt Le, Vishrav Chaudhary, Ahmed El-Kishky,
  Guillaume Wenzek, Myle Ott, and Marc'Aurelio Ranzato. 2019.
\newblock Facebook ai's wat19 myanmar-english translation task submission.
\newblock \emph{arXiv preprint arXiv:1910.06848}.

\bibitem[{Chen and Cardie(2018)}]{chen-cardie-2018-unsupervised}
Xilun Chen and Claire Cardie. 2018.
\newblock \href {https://doi.org/10.18653/v1/D18-1024} {Unsupervised
  multilingual word embeddings}.
\newblock In \emph{Proceedings of the 2018 Conference on Empirical Methods in
  Natural Language Processing}, pages 261--270, Brussels, Belgium. Association
  for Computational Linguistics.

\bibitem[{Chen et~al.(2017)Chen, Liu, Cheng, and Li}]{chen2017teacher}
Yun Chen, Yang Liu, Yong Cheng, and Victor~OK Li. 2017.
\newblock A teacher-student framework for zero-resource neural machine
  translation.
\newblock In \emph{Proceedings of the 55th Annual Meeting of the Association
  for Computational Linguistics (Volume 1: Long Papers)}, pages 1925--1935.

\bibitem[{Conneau et~al.(2019)Conneau, Khandelwal, Goyal, Chaudhary, Wenzek,
  Guzm{\'a}n, Grave, Ott, Zettlemoyer, and Stoyanov}]{conneau2019unsupervised}
Alexis Conneau, Kartikay Khandelwal, Naman Goyal, Vishrav Chaudhary, Guillaume
  Wenzek, Francisco Guzm{\'a}n, Edouard Grave, Myle Ott, Luke Zettlemoyer, and
  Veselin Stoyanov. 2019.
\newblock Unsupervised cross-lingual representation learning at scale.
\newblock \emph{arXiv preprint arXiv:1911.02116}.

\bibitem[{Conneau et~al.(2018)Conneau, Rinott, Lample, Williams, Bowman,
  Schwenk, and Stoyanov}]{conneau2018xnli}
Alexis Conneau, Ruty Rinott, Guillaume Lample, Adina Williams, Samuel~R.
  Bowman, Holger Schwenk, and Veselin Stoyanov. 2018.
\newblock Xnli: Evaluating cross-lingual sentence representations.
\newblock In \emph{Proceedings of the 2018 Conference on Empirical Methods in
  Natural Language Processing}. Association for Computational Linguistics.

\bibitem[{Devlin et~al.(2019)Devlin, Chang, Lee, and
  Toutanova}]{devlin2018bert}
Jacob Devlin, Ming-Wei Chang, Kenton Lee, and Kristina Toutanova. 2019.
\newblock {BERT}: Pre-training of deep bidirectional transformers for language
  understanding.
\newblock In \emph{North American Association for Computational Linguistics
  (NAACL)}.

\bibitem[{Ding et~al.(2019)Ding, {Hnin Thu Zar Aye}, {Win Pa Pa}, {Khin Thandar
  Nwet}, {Khin Mar Soe}, Utiyama, and Sumita}]{ding2019towards}
Chenchen Ding, {Hnin Thu Zar Aye}, {Win Pa Pa}, {Khin Thandar Nwet}, {Khin Mar
  Soe}, Masao Utiyama, and Eiichiro Sumita. 2019.
\newblock Towards {Burmese} ({Myanmar}) morphological analysis: Syllable-based
  tokenization and part-of-speech tagging.
\newblock \emph{ACM Transactions on Asian and Low-Resource Language Information
  Processing (TALLIP)}, 19(1):5.

\bibitem[{Ding et~al.(2018)Ding, Utiyama, and Sumita}]{ding2018nova}
Chenchen Ding, Masao Utiyama, and Eiichiro Sumita. 2018.
\newblock {NOVA}: A feasible and flexible annotation system for joint
  tokenization and part-of-speech tagging.
\newblock \emph{ACM Transactions on Asian and Low-Resource Language Information
  Processing (TALLIP)}, 18(2):17.

\bibitem[{Dong et~al.(2019)Dong, Yang, Wang, Wei, Liu, Wang, Gao, Zhou, and
  Hon}]{dong2019unified}
Li~Dong, Nan Yang, Wenhui Wang, Furu Wei, Xiaodong Liu, Yu~Wang, Jianfeng Gao,
  Ming Zhou, and Hsiao-Wuen Hon. 2019.
\newblock Unified language model pre-training for natural language
  understanding and generation.
\newblock \emph{arXiv preprint arXiv:1905.03197}.

\bibitem[{Edunov et~al.(2019)Edunov, Baevski, and Auli}]{edunov2019pre}
Sergey Edunov, Alexei Baevski, and Michael Auli. 2019.
\newblock Pre-trained language model representations for language generation.
\newblock \emph{arXiv preprint arXiv:1903.09722}.

\bibitem[{Firat et~al.(2016)Firat, Cho, and Bengio}]{firat2016multi}
Orhan Firat, Kyunghyun Cho, and Yoshua Bengio. 2016.
\newblock Multi-way, multilingual neural machine translation with a shared
  attention mechanism.
\newblock In \emph{NAACL}.

\bibitem[{Gu et~al.(2018)Gu, Hassan, Devlin, and Li}]{gu-etal-2018-universal}
Jiatao Gu, Hany Hassan, Jacob Devlin, and Victor~O.K. Li. 2018.
\newblock \href {https://doi.org/10.18653/v1/N18-1032} {Universal neural
  machine translation for extremely low resource languages}.
\newblock In \emph{Proceedings of the 2018 Conference of the North {A}merican
  Chapter of the Association for Computational Linguistics: Human Language
  Technologies, Volume 1 (Long Papers)}, pages 344--354, New Orleans,
  Louisiana. Association for Computational Linguistics.

\bibitem[{Gu et~al.(2019)Gu, Wang, Cho, and Li}]{gu2019improved}
Jiatao Gu, Yong Wang, Kyunghyun Cho, and Victor~OK Li. 2019.
\newblock Improved zero-shot neural machine translation via ignoring spurious
  correlations.
\newblock \emph{arXiv preprint arXiv:1906.01181}.

\bibitem[{Guzm{\'a}n et~al.(2019)Guzm{\'a}n, Chen, Ott, Pino, Lample, Koehn,
  Chaudhary, and Ranzato}]{guzman-etal-2019-flores}
Francisco Guzm{\'a}n, Peng-Jen Chen, Myle Ott, Juan Pino, Guillaume Lample,
  Philipp Koehn, Vishrav Chaudhary, and Marc{'}Aurelio Ranzato. 2019.
\newblock \href {https://doi.org/10.18653/v1/D19-1632} {The {FLORES} evaluation
  datasets for low-resource machine translation: {N}epali{--}{E}nglish and
  {S}inhala{--}{E}nglish}.
\newblock In \emph{Proceedings of the 2019 Conference on Empirical Methods in
  Natural Language Processing and the 9th International Joint Conference on
  Natural Language Processing (EMNLP-IJCNLP)}, pages 6097--6110, Hong Kong,
  China. Association for Computational Linguistics.

\bibitem[{Jean et~al.(2017)Jean, Lauly, Firat, and
  Cho}]{DBLP:journals/corr/JeanLFC17}
S{\'{e}}bastien Jean, Stanislas Lauly, Orhan Firat, and Kyunghyun Cho. 2017.
\newblock \href {http://arxiv.org/abs/1704.05135} {Does neural machine
  translation benefit from larger context?}
\newblock \emph{CoRR}, abs/1704.05135.

\bibitem[{Johnson et~al.(2017)Johnson, Schuster, Le, Krikun, Wu, Chen, Thorat,
  Vi{\'e}gas, Wattenberg, Corrado et~al.}]{johnson2017google}
Melvin Johnson, Mike Schuster, Quoc~V Le, Maxim Krikun, Yonghui Wu, Zhifeng
  Chen, Nikhil Thorat, Fernanda Vi{\'e}gas, Martin Wattenberg, Greg Corrado,
  et~al. 2017.
\newblock Google’s multilingual neural machine translation system: Enabling
  zero-shot translation.
\newblock \emph{Transactions of the Association for Computational Linguistics},
  5:339--351.

\bibitem[{Kudo and Richardson(2018)}]{kudo-richardson-2018-sentencepiece}
Taku Kudo and John Richardson. 2018.
\newblock \href {https://doi.org/10.18653/v1/D18-2012} {{S}entence{P}iece: A
  simple and language independent subword tokenizer and detokenizer for neural
  text processing}.
\newblock In \emph{Proceedings of the 2018 Conference on Empirical Methods in
  Natural Language Processing: System Demonstrations}, pages 66--71, Brussels,
  Belgium. Association for Computational Linguistics.

\bibitem[{Kunchukuttan et~al.(2017)Kunchukuttan, Mehta, and
  Bhattacharyya}]{DBLP:journals/corr/abs-1710-02855}
Anoop Kunchukuttan, Pratik Mehta, and Pushpak Bhattacharyya. 2017.
\newblock \href {http://arxiv.org/abs/1710.02855} {The {IIT} bombay
  english-hindi parallel corpus}.
\newblock \emph{CoRR}, abs/1710.02855.

\bibitem[{Lample and Conneau(2019)}]{lample2019cross}
Guillaume Lample and Alexis Conneau. 2019.
\newblock Cross-lingual language model pretraining.
\newblock \emph{arXiv preprint arXiv:1901.07291}.

\bibitem[{Lample et~al.(2018{\natexlab{a}})Lample, Conneau, Denoyer, and
  Ranzato}]{lample2018unsupervised}
Guillaume Lample, Alexis Conneau, Ludovic Denoyer, and Marc'Aurelio Ranzato.
  2018{\natexlab{a}}.
\newblock \href {https://openreview.net/forum?id=rkYTTf-AZ} {Unsupervised
  machine translation using monolingual corpora only}.
\newblock In \emph{International Conference on Learning Representations}.

\bibitem[{Lample et~al.(2018{\natexlab{b}})Lample, Conneau, Ranzato, Denoyer,
  and Jégou}]{lample2018word}
Guillaume Lample, Alexis Conneau, Marc'Aurelio Ranzato, Ludovic Denoyer, and
  Hervé Jégou. 2018{\natexlab{b}}.
\newblock \href {https://openreview.net/forum?id=H196sainb} {Word translation
  without parallel data}.
\newblock In \emph{International Conference on Learning Representations}.

\bibitem[{Lample et~al.(2018{\natexlab{c}})Lample, Ott, Conneau, Denoyer, and
  Ranzato}]{lample2018phrase}
Guillaume Lample, Myle Ott, Alexis Conneau, Ludovic Denoyer, and Marc'Aurelio
  Ranzato. 2018{\natexlab{c}}.
\newblock Phrase-based \& neural unsupervised machine translation.
\newblock \emph{arXiv preprint arXiv:1804.07755}.

\bibitem[{Lewis et~al.(2019)Lewis, Liu, Goyal, Ghazvininejad, Mohamed, Levy,
  Stoyanov, and Zettlemoyer}]{lewis2019bart}
Mike Lewis, Yinhan Liu, Naman Goyal, Marjan Ghazvininejad, Abdelrahman Mohamed,
  Omer Levy, Veselin Stoyanov, and Luke Zettlemoyer. 2019.
\newblock Bart: Denoising sequence-to-sequence pre-training for natural
  language generation, translation, and comprehension.
\newblock \emph{arXiv preprint arXiv:1910.13461}.

\bibitem[{Li et~al.(2019)Li, Jiang, and Liu}]{li2019pretrained}
Liangyou Li, Xin Jiang, and Qun Liu. 2019.
\newblock Pretrained language models for document-level neural machine
  translation.
\newblock \emph{arXiv preprint arXiv:1911.03110}.

\bibitem[{Liu and Lapata(2019)}]{liu2019text}
Yang Liu and Mirella Lapata. 2019.
\newblock Text summarization with pretrained encoders.
\newblock \emph{arXiv preprint arXiv:1908.08345}.

\bibitem[{Liu et~al.(2019)Liu, Ott, Goyal, Du, Joshi, Chen, Levy, Lewis,
  Zettlemoyer, and Stoyanov}]{liu2019roberta}
Yinhan Liu, Myle Ott, Naman Goyal, Jingfei Du, Mandar Joshi, Danqi Chen, Omer
  Levy, Mike Lewis, Luke Zettlemoyer, and Veselin Stoyanov. 2019.
\newblock Roberta: A robustly optimized bert pretraining approach.
\newblock \emph{arXiv preprint arXiv:1907.11692}.

\bibitem[{Miculicich et~al.(2018)Miculicich, Ram, Pappas, and
  Henderson}]{miculicich-etal-2018-document}
Lesly Miculicich, Dhananjay Ram, Nikolaos Pappas, and James Henderson. 2018.
\newblock \href {https://doi.org/10.18653/v1/D18-1325} {Document-level neural
  machine translation with hierarchical attention networks}.
\newblock In \emph{Proceedings of the 2018 Conference on Empirical Methods in
  Natural Language Processing}, pages 2947--2954, Brussels, Belgium.
  Association for Computational Linguistics.

\bibitem[{Mikolov et~al.(2013)Mikolov, Le, and
  Sutskever}]{DBLP:journals/corr/MikolovLS13}
Tomas Mikolov, Quoc~V. Le, and Ilya Sutskever. 2013.
\newblock \href {http://arxiv.org/abs/1309.4168} {Exploiting similarities among
  languages for machine translation}.
\newblock \emph{CoRR}, abs/1309.4168.

\bibitem[{Ott et~al.(2019)Ott, Edunov, Baevski, Fan, Gross, Ng, Grangier, and
  Auli}]{ott2019fairseq}
Myle Ott, Sergey Edunov, Alexei Baevski, Angela Fan, Sam Gross, Nathan Ng,
  David Grangier, and Michael Auli. 2019.
\newblock \textsc{fairseq}: A fast, extensible toolkit for sequence modeling.
\newblock In \emph{North American Association for Computational Linguistics
  (NAACL): System Demonstrations}.

\bibitem[{Papineni et~al.(2002)Papineni, Roukos, Ward, and
  Zhu}]{papineni2002bleu}
Kishore Papineni, Salim Roukos, Todd Ward, and Wei-Jing Zhu. 2002.
\newblock Bleu: a method for automatic evaluation of machine translation.
\newblock In \emph{Proceedings of the 40th annual meeting on association for
  computational linguistics}, pages 311--318. Association for Computational
  Linguistics.

\bibitem[{Peters et~al.(2018)Peters, Neumann, Iyyer, Gardner, Clark, Lee, and
  Zettlemoyer}]{peters2018deep}
Matthew Peters, Mark Neumann, Mohit Iyyer, Matt Gardner, Christopher Clark,
  Kenton Lee, and Luke Zettlemoyer. 2018.
\newblock Deep contextualized word representations.
\newblock In \emph{North American Association for Computational Linguistics
  (NAACL)}.

\bibitem[{Post(2018)}]{post-2018-call}
Matt Post. 2018.
\newblock \href {https://www.aclweb.org/anthology/W18-6319} {A call for clarity
  in reporting {BLEU} scores}.
\newblock In \emph{Proceedings of the Third Conference on Machine Translation:
  Research Papers}, pages 186--191, Belgium, Brussels. Association for
  Computational Linguistics.

\bibitem[{Pourdamghani et~al.(2019)Pourdamghani, Aldarrab, Ghazvininejad,
  Knight, and May}]{Pourdamghani2019TranslatingTA}
Nima Pourdamghani, Nada Aldarrab, Marjan Ghazvininejad, Kevin Knight, and
  Jonathan May. 2019.
\newblock Translating translationese: A two-step approach to unsupervised
  machine translation.
\newblock In \emph{ACL}.

\bibitem[{Radford et~al.(2018)Radford, Narasimhan, Salimans, and
  Sutskever}]{radford2018gpt}
Alec Radford, Karthik Narasimhan, Time Salimans, and Ilya Sutskever. 2018.
\newblock Improving language understanding with unsupervised learning.
\newblock Technical report, OpenAI.

\bibitem[{Radford et~al.(2019)Radford, Wu, Child, Luan, Amodei, and
  Sutskever}]{radford2019language}
Alec Radford, Jeffrey Wu, Rewon Child, David Luan, Dario Amodei, and Ilya
  Sutskever. 2019.
\newblock Language models are unsupervised multitask learners.
\newblock Technical report, OpenAI.

\bibitem[{Raffel et~al.(2019)Raffel, Shazeer, Roberts, Lee, Narang, Matena,
  Zhou, Li, and Liu}]{raffel2019exploring}
Colin Raffel, Noam Shazeer, Adam Roberts, Katherine Lee, Sharan Narang, Michael
  Matena, Yanqi Zhou, Wei Li, and Peter~J Liu. 2019.
\newblock Exploring the limits of transfer learning with a unified text-to-text
  transformer.
\newblock \emph{arXiv preprint arXiv:1910.10683}.

\bibitem[{Sennrich et~al.(2016{\natexlab{a}})Sennrich, Haddow, and
  Birch}]{sennrich2016edinburgh}
Rico Sennrich, Barry Haddow, and Alexandra Birch. 2016{\natexlab{a}}.
\newblock Edinburgh neural machine translation systems for wmt 16.
\newblock In \emph{Proceedings of the First Conference on Machine Translation:
  Volume 2, Shared Task Papers}, pages 371--376.

\bibitem[{Sennrich et~al.(2016{\natexlab{b}})Sennrich, Haddow, and
  Birch}]{sennrich-etal-2016-improving}
Rico Sennrich, Barry Haddow, and Alexandra Birch. 2016{\natexlab{b}}.
\newblock \href {https://doi.org/10.18653/v1/P16-1009} {Improving neural
  machine translation models with monolingual data}.
\newblock In \emph{Proceedings of the 54th Annual Meeting of the Association
  for Computational Linguistics (Volume 1: Long Papers)}, pages 86--96, Berlin,
  Germany. Association for Computational Linguistics.

\bibitem[{Shirish~Keskar et~al.(2019)Shirish~Keskar, McCann, Varshney, Xiong,
  and Socher}]{shirish2019ctrl}
Nitish Shirish~Keskar, Bryan McCann, Lav~R Varshney, Caiming Xiong, and Richard
  Socher. 2019.
\newblock Ctrl: A conditional transformer language model for controllable
  generation.
\newblock \emph{arXiv preprint arXiv:1909.05858}.

\bibitem[{Song et~al.(2019)Song, Tan, Qin, Lu, and Liu}]{song2019mass}
Kaitao Song, Xu~Tan, Tao Qin, Jianfeng Lu, and Tie-Yan Liu. 2019.
\newblock {MASS}: Masked sequence to sequence pre-training for language
  generation.
\newblock In \emph{International Conference on Machine Learning (ICML)}.

\bibitem[{Tiedemann and Scherrer(2017)}]{tiedemann-scherrer-2017-neural}
J{\"o}rg Tiedemann and Yves Scherrer. 2017.
\newblock \href {https://doi.org/10.18653/v1/W17-4811} {Neural machine
  translation with extended context}.
\newblock In \emph{Proceedings of the Third Workshop on Discourse in Machine
  Translation}, pages 82--92, Copenhagen, Denmark. Association for
  Computational Linguistics.

\bibitem[{Tu et~al.(2018)Tu, Liu, Shi, and Zhang}]{doi:10.1162/tacla00029}
Zhaopeng Tu, Yang Liu, Shuming Shi, and Tong Zhang. 2018.
\newblock \href {https://doi.org/10.1162/tacl\_a\_00029} {Learning to remember
  translation history with a continuous cache}.
\newblock \emph{Transactions of the Association for Computational Linguistics},
  6:407--420.

\bibitem[{Vaswani et~al.(2017)Vaswani, Shazeer, Parmar, Uszkoreit, Jones,
  Gomez, Kaiser, and Polosukhin}]{vaswani2017attention}
Ashish Vaswani, Noam Shazeer, Niki Parmar, Jakob Uszkoreit, Llion Jones,
  Aidan~N Gomez, {\L}ukasz Kaiser, and Illia Polosukhin. 2017.
\newblock Attention is all you need.
\newblock In \emph{Advances in neural information processing systems}.

\bibitem[{Vi{\'e}gas et~al.(2016)Vi{\'e}gas, Corrado, Dean, Hughes, Wattenberg,
  Krikun, Johnson, Schuster, Thorat, Le et~al.}]{viegas2016google}
Fernanda Vi{\'e}gas, Greg Corrado, Jeffrey Dean, Macduff Hughes, Martin
  Wattenberg, Maxim Krikun, Melvin Johnson, Mike Schuster, Nikhil Thorat,
  Quoc~V Le, et~al. 2016.
\newblock Google's multilingual neural machine translation system: Enabling
  zero-shot translation.

\bibitem[{Wada and Iwata(2018)}]{DBLP:wada}
Takashi Wada and Tomoharu Iwata. 2018.
\newblock \href {http://arxiv.org/abs/1809.02306} {Unsupervised cross-lingual
  word embedding by multilingual neural language models}.
\newblock \emph{CoRR}, abs/1809.02306.

\bibitem[{Wang et~al.(2017)Wang, Tu, Way, and
  Liu}]{wang-etal-2017-exploiting-cross}
Longyue Wang, Zhaopeng Tu, Andy Way, and Qun Liu. 2017.
\newblock \href {https://doi.org/10.18653/v1/D17-1301} {Exploiting
  cross-sentence context for neural machine translation}.
\newblock In \emph{Proceedings of the 2017 Conference on Empirical Methods in
  Natural Language Processing}, pages 2826--2831, Copenhagen, Denmark.
  Association for Computational Linguistics.

\bibitem[{Wenzek et~al.(2019)Wenzek, Lachaux, Conneau, Chaudhary, Guzman,
  Joulin, and Grave}]{wenzek2019ccnet}
Guillaume Wenzek, Marie-Anne Lachaux, Alexis Conneau, Vishrav Chaudhary,
  Francisco Guzman, Armand Joulin, and Edouard Grave. 2019.
\newblock Ccnet: Extracting high quality monolingual datasets from web crawl
  data.
\newblock \emph{arXiv preprint arXiv:1911.00359}.

\bibitem[{Wu et~al.(2019{\natexlab{a}})Wu, Wang, and Wang}]{wu2019extract}
Jiawei Wu, Xin Wang, and William~Yang Wang. 2019{\natexlab{a}}.
\newblock Extract and edit: An alternative to back-translation for unsupervised
  neural machine translation.
\newblock \emph{arXiv preprint arXiv:1904.02331}.

\bibitem[{Wu et~al.(2019{\natexlab{b}})Wu, Zhu, He, Gao, Tan, Qin, and
  Liu}]{wu2019machine}
Lijun Wu, Jinhua Zhu, Di~He, Fei Gao, Xu~Tan, Tao Qin, and Tie-Yan Liu.
  2019{\natexlab{b}}.
\newblock \href {https://openreview.net/forum?id=ryza73R9tQ} {Machine
  translation with weakly paired bilingual documents}.

\bibitem[{Yang et~al.(2019)Yang, Dai, Yang, Carbonell, Salakhutdinov, and
  Le}]{yang2019xlnet}
Zhilin Yang, Zihang Dai, Yiming Yang, Jaime Carbonell, Ruslan Salakhutdinov,
  and Quoc~V Le. 2019.
\newblock Xlnet: Generalized autoregressive pretraining for language
  understanding.
\newblock \emph{arXiv preprint arXiv:1906.08237}.

\bibitem[{Zhang et~al.(2019)Zhang, Sun, Galley, Chen, Brockett, Gao, Gao, Liu,
  and Dolan}]{zhang2019dialogpt}
Yizhe Zhang, Siqi Sun, Michel Galley, Yen-Chun Chen, Chris Brockett, Xiang Gao,
  Jianfeng Gao, Jingjing Liu, and Bill Dolan. 2019.
\newblock \href {http://arxiv.org/abs/1911.00536} {Dialogpt: Large-scale
  generative pre-training for conversational response generation}.

\end{thebibliography}
\appendix
\section{Evaluation Details}
\label{sec:appendix}
For all our tasks, we use BLEU scores~\cite{papineni2002bleu} as the automatic metric to evaluate the translation performance. Normally, we compute the BLEU scores over tokenized text for both system outputs and the references, and we apply language-wise tokenization after over the translation. Note that, since we directly work on raw texts, we automatically get de-tokenized output after recovering sentence-piece subwords. Following the literature, the instructions of language-wise tokenization are as follows:
\begin{itemize}[leftmargin=*]
    \item {\bf Gu, Ne, Si, Hi}: We use Indic-NLP Library~\footnote{\url{https://anoopkunchukuttan.github.io/indic_nlp_library/}} to tokenize the Indic language outputs.
    \item {\bf Ja}: We use KyTea~\footnote{\url{http://www.phontron.com/kytea/}} to segment Japanese texts.
    \item {\bf Ko}: We use Mecab-Ko~\footnote{\url{http://konlpy.org/en/v0.3.0/install/}} and its default dictionary to segment the Korean texts 
    \item {\bf Ar}: We apply QCRI Arabic Normalizer~\footnote{\url{http://alt.qcri.org/tools/arabic-normalizer/}} over the Arabic texts.
    \item {\bf My}: We use the official segmentation tool provided by~\newcite{ding2019towards} for Burmese.
    \item {\bf Ro}: Following \newcite{sennrich2016edinburgh}, we apply Moses tokenization and special normalization for Romanian texts~\footnote{\url{https://github.com/rsennrich/wmt16-script}}.
    \item {\bf Zh}: We use the official sacreBleu~\cite{post-2018-call}\footnote{\url{https://github.com/mjpost/sacreBLEU}} Chinese tokenizer (--tok zh). 
    
\end{itemize}
For other languages that are not listed above, we compute BLEU scores with sacreBLEU with DEFAULT tokenization.

\section{Translation Examples}
\label{sec:example}
\begin{figure*}[t]
    \centering
    \includegraphics[width=\linewidth]{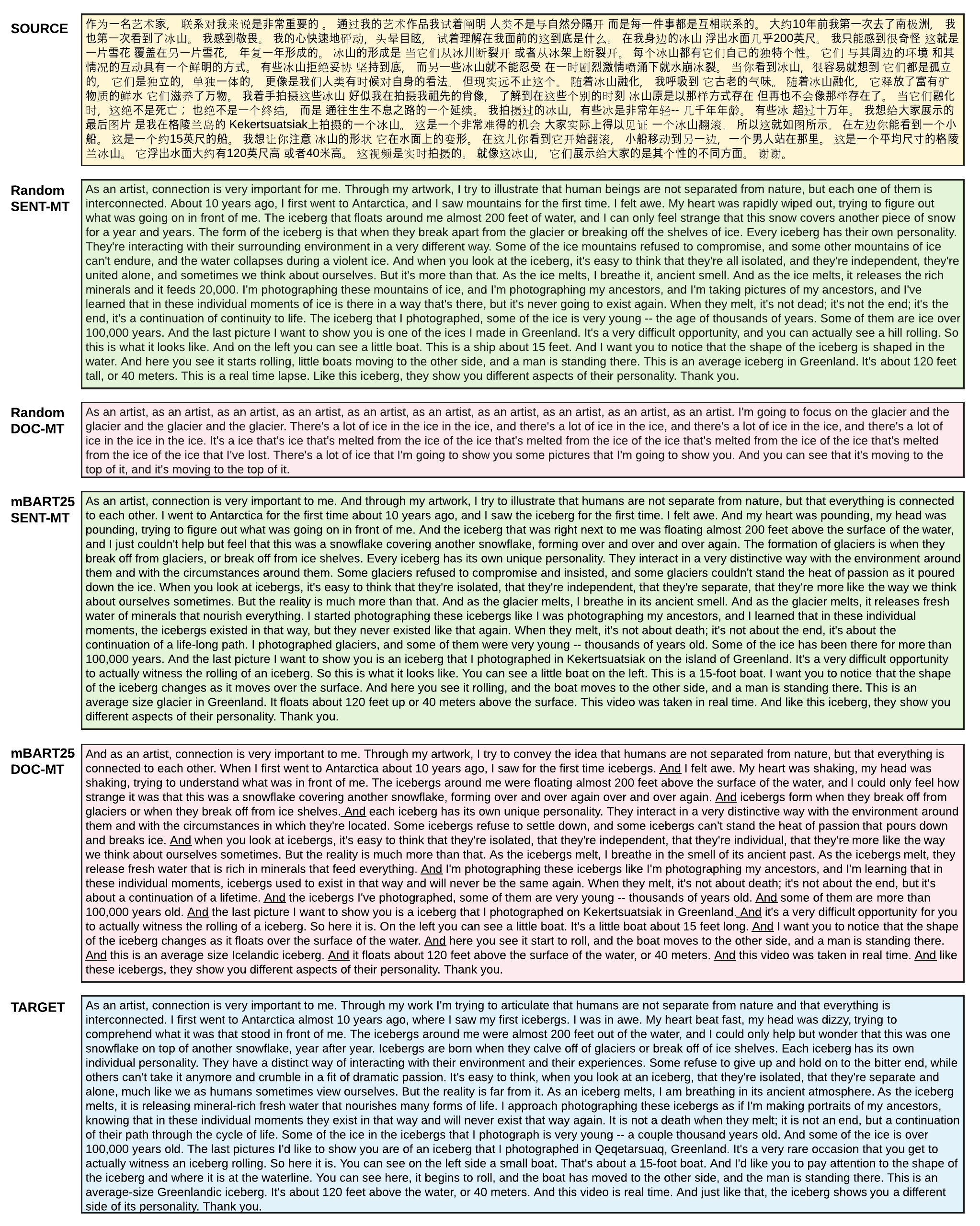}
    \caption{{\bf An Example of Document-level translation from mBART25 Sent-MT and Doc-MT}, held out from the test set of TED15 Zh-En. The Doc-MT system produces much fluent and coherent translation which is closer to the reference translation. For instance, Doc-MT model produces several ``\underline{And}'' to connect sentences to make it reads better, while the Sent-MT model does not contain global knowledge and produce sentences independently. Besides, both systems produce much better translations than models without pre-training where the non-pretrained Doc-MT model completely fails to produce readable translation output.}
    \label{fig:example_doc}
\end{figure*}

\begin{figure*}
    \centering
    \includegraphics[width=0.95\linewidth]{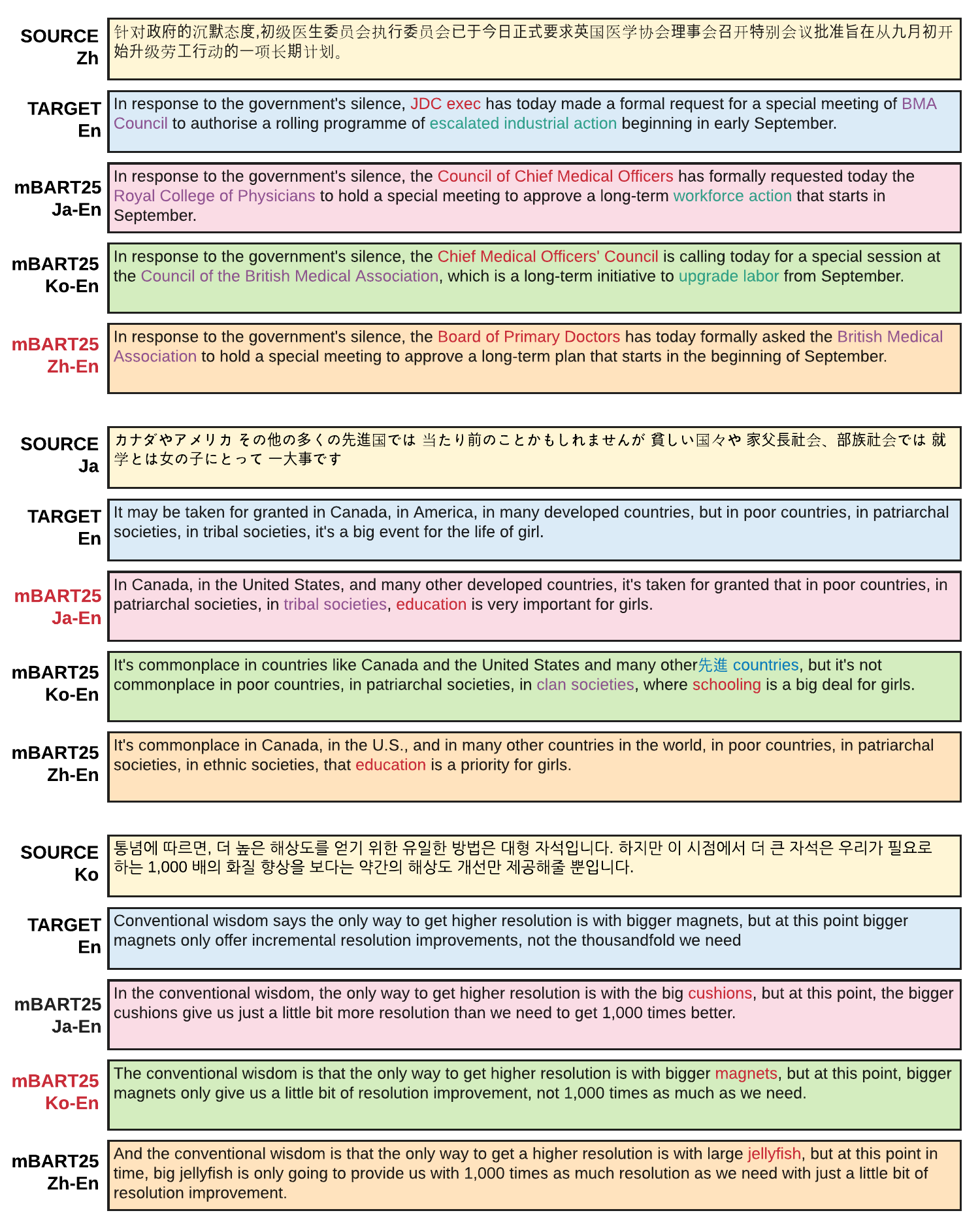}
    \caption{{\bf Examples of Unsupervised MT via Language Transfer} between {\bf Ja}, {\bf Ko}, {\bf Zh} $\rightarrow$ {\bf En}. We mark the supervised settings in {\color{red} red}. All three languages have quite different character sets (Ja and Zh shares part of the Chinese characters) and syntactic structures. However, they are still culturally and historically correlated, which we assume can be captured through pre-training. For all cases, if we fine-tune the mBART25 model on any pair, the resulted model directly translates well in the other two pairs without seeing any corresponded parallel sentences. We also see failure cases. For instance (the 3rd example), only the supervised model translates ``자석'' into ``magents'' correctly, while the Ja-En and Zh-En guess with irreverent words ``cushions'' and ``jellyfish'', respectively. Also, in the 2nd example, the Ko-En model fails to translate ``developed'' and copies the source tokens. We suspect it is because the pre-training stage biases the output distribution. }
    \label{fig:my_label}
\end{figure*}
\end{document}